\documentclass[10pt,journal,compsoc]{IEEEtran}

\usepackage[utf8]{inputenc} %
\usepackage[T1]{fontenc}    %
\usepackage{hyperref}       %
\usepackage{url}            %
\usepackage{booktabs}       %
\usepackage{amsfonts}       %
\usepackage{bbm}
\usepackage{nicefrac}       %
\usepackage{microtype}      %
\usepackage[normalem]{ulem}

\ifCLASSOPTIONcompsoc
  \usepackage[nocompress]{cite}
\else
  \usepackage{cite}
\fi

\usepackage{graphicx}
\usepackage{amsmath,amssymb} %
\usepackage{times}
\usepackage{epsfig}
\usepackage{amsmath}
\usepackage{amssymb}
\usepackage{wrapfig}
\usepackage[export]{adjustbox}
\usepackage{tikz}
\usepackage{color}
\usepackage{enumitem}
\usepackage{hyperref}
\usepackage{thmtools}
\usepackage{thm-restate}
\usepackage{soul}
\usepackage{dsfont}

\DeclareMathOperator*{\argmin}{arg\,min}
\usepackage{mathrsfs}
\newcommand*{\subproblem}{S}%
\newcommand*{\Ainit}{\tilde{A}}%
\newcommand*{\intSC}{ISC}%
\newcommand*{\gt}[1]{\stackrel{\smash{*}\rule{0pt}{-1.ex}}{#1}}%
\newcommand{\RED}[1]{#1}

\definecolor{cycolor}{HTML}{00F9DE}
\newcommand\REVIEW[1]{#1}
\newcommand\operator[1]{\textrm{#1}}

\usepackage[]{algorithm2e}

\usepackage{epstopdf}
\DeclareGraphicsExtensions{.eps}

\usepackage{subcaption}

\usepackage{booktabs}

\usepackage{algpseudocode}

\usepackage{amsthm}
\newtheorem{theorem}{Theorem}[section]

\newtheorem{lemma}[theorem]{Lemma}

\newtheorem{definition}{Definition}[section]

\begin{document}

\title{The Mutex Watershed and its Objective: \\ Efficient, Parameter-Free Signed Graph Partitioning}
\title{The Mutex Watershed and its Objective:\\ Efficient, Parameter-Free Graph Partitioning}

\author{Steffen~Wolf$^*$\thanks{$^*$ Authors contributed equally}, %
        Alberto~Bailoni$^*$, %
        Constantin~Pape, %
        Nasim~Rahaman, \\
        Anna~Kreshuk, %
        Ullrich~K\"othe, %
        and Fred A. Hamprecht$^\dagger$\thanks{$^\dagger$ Corresponding author}%
\IEEEcompsocitemizethanks{%
\IEEEcompsocthanksitem All authors are with HCI/IWR, Heidelberg University, Germany.\protect\\
E-mail: <firstname>.<lastname>@iwr.uni-heidelberg.de
\IEEEcompsocthanksitem A.~Kreshuk and C.~Pape are with EMBL, Heidelberg, Germany.}%
}%

\markboth{IEEE TRANSACTIONS ON PATTERN ANALYSIS AND MACHINE INTELLIGENCE}%
{Shell \MakeLowercase{\textit{et al.}}: Bare Demo of IEEEtran.cls for Computer Society Journals}

\IEEEtitleabstractindextext{%
\begin{abstract}
Image partitioning, or segmentation without semantics, is the task of decomposing an image into distinct segments, or equivalently to detect closed contours. Most prior work either requires seeds, one per segment; or a threshold; or formulates the task as multicut / correlation clustering, an NP-hard problem. Here, we propose an \REVIEW{efficient} algorithm for \REVIEW{graph partitioning}, the ``Mutex Watershed''. Unlike seeded watershed, the algorithm can accommodate not only attractive but also repulsive cues, allowing it to find a previously \emph{unspecified} number of segments without the need for explicit seeds or a tunable threshold. We also prove that this simple algorithm solves to global optimality an objective function that is intimately related to the multicut / correlation clustering integer linear programming formulation. 
The algorithm is deterministic, very simple to implement, and has empirically linearithmic complexity. 
When presented with short-range attractive and long-range repulsive cues from a deep neural network, the Mutex Watershed gives the best results currently known for the %
competitive ISBI 2012 EM segmentation benchmark. %
\end{abstract}

}

\maketitle

\IEEEdisplaynontitleabstractindextext

\IEEEpeerreviewmaketitle

\IEEEraisesectionheading{\section{Introduction}\label{intro}}

\noindent \IEEEPARstart{M}{ost} image partitioning algorithms are defined over a graph encoding purely attractive interactions. No matter whether a segmentation or clustering is then found agglomeratively (as in single linkage clustering / watershed) or divisively (as in spectral clustering or iterated normalized cuts), the user either needs to specify the desired number of segments or a termination criterion. An even stronger form of supervision is in terms of seeds, where one pixel of each segment needs to be designated either by a user or automatically. Unfortunately, clustering with 
automated seed selection remains a fragile and error-fraught process, because every missed or hallucinated seed causes an under- or oversegmentation error. Although the learning of good edge detectors boosts the quality of classical seed selection strategies (such as finding local minima of the boundary map, or thresholding boundary maps), non-local effects of seed placement along with strong variability in region sizes and shapes make it hard for any learned predictor to place {\em exactly one} seed in every true region.

In contrast to the above class of algorithms, multicut / correlation clustering partitions vertices with both attractive and repulsive interactions encoded into the edges of a graph. Multicut has the great advantage that a ``natural'' partitioning of a graph can be found, without needing to specify a  desired number of clusters, or a termination criterion, or one seed per region. Its great drawback is that its optimization is NP-hard. 

\begin{figure}[t]
    \centering
    \includegraphics[width=1.\linewidth,valign=t]{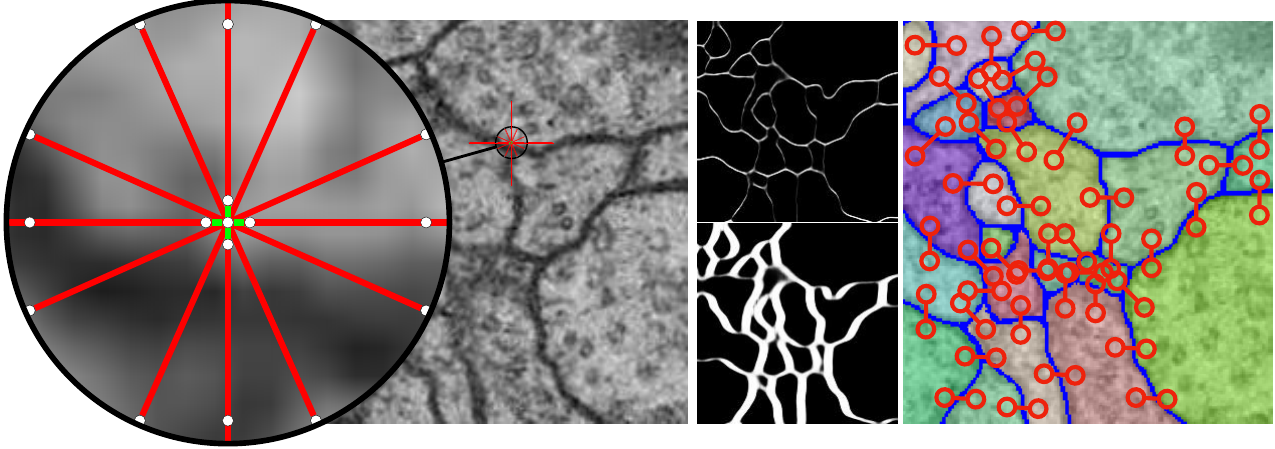}%
    \caption{Left: Overlay of raw data from the ISBI 2012 EM segmentation challenge and the edges for which attractive (green) or repulsive (red) interactions are estimated for each pixel using a CNN. Middle: vertical / horizontal repulsive interactions at intermediate / long range are shown in the top / bottom half. Right: Active mutual exclusion (mutex) constraints that the proposed algorithm invokes during the segmentation process.}
    \label{fig:main}
    \vspace{-0.05cm}
\end{figure}

The main insight of this paper is that when both attractive and repulsive interactions between pixels are available, then a generalization of the watershed algorithm can be devised that segments an image {\em without} the need for seeds or stopping criteria or thresholds. It examines all graph edges, attractive and repulsive, sorted by their weight and adds these to an active set iff they are not in conflict with previous, higher-priority, decisions. The attractive subset of the resulting active set  is a forest, with one tree representing each segment. However, the active set can have loops involving more than one repulsive edge.
 See Fig.~\ref{fig:main} for a visual abstract. 

In summary, our principal contributions are, first, 
a fast deterministic algorithm for \REVIEW{graph partitioning with both positive and negative edge weights} that does not need prior specification of the number of clusters~(section \ref{sec:MWS_objective}); and second, its theoretical characterization, including proof that it globally optimizes an objective related to the multicut correlation clustering objective~(\ref{sec:MWS_objective}).

Combined with a deep net, the algorithm also happens to define the state-of-the-art in a competitive neuron segmentation challenge~(\autoref{4_results}).

This is an extended version version of \cite{wolf2018mutex}, with the second principal contribution (section \ref{sec:MWS_objective}) being new.

\section{Related Work} \label{2_rel_work}

\noindent In the original watershed algorithm \cite{vincent1991watersheds,Beucher-Lantu-79}, seeds were automatically placed at all local minima of the boundary map. Unfortunately, this leads to severe over-segmentation. Defining better seeds has been a recurring theme of watershed research ever since. The simplest solution is offered by the seeded watershed algorithm \cite{beucher1992morphological}: It relies on an oracle (an external algorithm or a human) to provide seeds and assigns each pixel to its nearest seed in terms of minimax path distance.

\REVIEW{In the absence of an oracle, many automatic methods for seed selection have been proposed in the last decades with applications in the fields of medicine and biology. Many of these approaches rely on edge feature extraction and edge detection like gradient calculation \cite{pohle2001segmentation,alattar2010myocardial}. Other types of methods generate seeds by first performing feature extraction \cite{poonguzhali2006complete,wu2008texture}, whereas others first extract region of interests and then place seeds inside these regions by using thresholding \cite{al2014computer}, binarization \cite{shan2008novel}, $k$-means \cite{mubarak2012hybrid} or other strategies \cite{abdelsamea2011enhancement,al2014breast}. 
}

\REVIEW{In applications where the number of regions is hard to estimate, simple automatic seed selection methods, e.g.\ defining seeds by connected regions of low boundary probability, don't work: The segmentation quality is usually insufficient because multiple seeds are in the same region and/or seeds leak through the boundary.} Thus, in these cases seed selection may be biased towards over-seg\-men\-ta\-tion (with seeding at all minima being the extreme case). The watershed algorithm then produces superpixels that are merged into final regions by more or less elaborate postprocessing. This works better than using watersheds alone because it exploits the larger context afforded by superpixel adjacency graphs. Many criteria have been proposed to identify the regions to be preserved during merging, e.g.\ region dynamics \cite{grimaud_92_watershed-dynamics}, the waterfall transform \cite{beucher1994watershed}, extinction values \cite{vachier1995extinction}, region saliency \cite{najman1996geodesic}, and $(\alpha,\omega)$-connected components \cite{soille_08_hierarchical-image-decomposition}. A merging process controlled by criteria like these can be iterated to produce a hierarchy of segmentations where important regions survive to the next level. Variants of such hierarchical watersheds are reviewed and evaluated in \cite{perret_17_hierarchical-watersheds}.

These results highlight the close connection of watersheds to hierarchical clustering and minimum spanning trees/forests \cite{meyer1999morphological,najman_11_ultrametric-watersheds}, which inspired novel merging strategies and termination criteria. For example, \cite{salembier_00_binary-partition-tree} simply terminated hierarchical merging by fixing the number of surviving regions beforehand. \cite{malmberg2011generalized} incorporate predefined sets of generalized merge constraints into the clustering algorithm. Graph-based segmentation according to \cite{felzenszwalb_04_graph-based-image-segmentation} defines a measure of quality for the current regions and stops when the merge costs would exceed this measure. Ultrametric contour maps \cite{arbelaez_11_gpb} combine the gPb (global probability of boundary) edge detector with an oriented watershed transform. Superpixels are agglomerated until the ultrametric distance between the resulting regions exceeds a learned threshold. An optimization perspective is taken in \cite{kiran_13_hierarchical-cuts,guigues2006scale}, which introduces $h$-increasing energy functions and builds the hierarchy incrementally such that merge decisions greedily minimize the energy. The authors prove that the optimal cut corresponds to a different unique segmentation for every value of a free regularization parameter.

\REVIEW{An important line of research is given by partitioning of graphs with both attractive and repulsive edges \cite{keuper2016multi}.} Solutions that optimally balance attraction and repulsion do not require external stopping criteria such as predefined number of regions or seeds. This generalization leads to the NP-hard problem of correlation clustering or (synonymous) multicut (MC) partitioning. Fortunately, modern integer linear programming solvers in combination with incremental constraint generation can solve problem instances of considerable size \cite{andres_12_globally}, and good approximations exist for even larger problems \cite{yarkony2012fast,pape2017solving}
Reminiscent of strict minimizers \cite{Levi} with minimal $L_\infty$-norm solution, our work solves the multicut objective optimally when all graph weights are raised to a large power.

Related to the proposed method, the greedy additive edge contraction (GAEC) \cite{keuper2015efficient} heuristic for the multicut also sequentially merges regions, but we handle attractive and repulsive interactions separately and define edge strength between clusters by a maximum instead of an additive rule.
The greedy fixation algorithm introduced in \cite{levinkov2017comparative} is closely related to the proposed method; it sorts attractive and repulsive edges by their absolute
weight, merges nodes connected by attractive edges and introduces no-merge constraints
for repulsive edges. However, similar to GAEC, it defines edge strength by an additive rule,
which increases the algorithm's runtime complexity compared to the presented Mutex Watershed. Also, it is not yet known what objective the algorithm optimizes globally, if any.

Another beneficial extension is the introduction of additional long-range edges. The strength of such edges can often be estimated with greater certainty than is achievable for the local edges used by watersheds on standard 4- or 8-connected pixel graphs. Such repulsive long-range edges have been used in \cite{zhang_14_cell} to represent object diameter constraints, which is still an MC-type problem. When long-range edges are also allowed to be attractive, the problem turns into the more complicated lifted multicut (LMC) \cite{horvnakova2017analysis}. Realistic problem sizes can only be solved approximately \cite{keuper2015efficient,beier2016efficient}, but watershed superpixels followed by LMC postprocessing achieve state-of-the-art results on important benchmarks \cite{beier2017multicut}. Long-range edges are also used in \cite{lee2017superhuman}, as side losses for the boundary detection convolutional neural network~(CNN); but they are not used explicitly in any downstream inference.

In general, striking progress in watershed-based segmentation has been achieved by learning boundary maps with CNNs. This is nicely illustrated by the evolution of neurosegmentation for connectomics, an important field we also address in the experimental section. CNNs were introduced to this application in \cite{jain2007supervised} and became, in much refined form \cite{ciresan_12_deep-em-segmentation}, the winning entry of the ISBI 2012 Neuro-Segmentation Challenge \cite{isbi2012challenge}. Boundary maps and superpixels were further improved by progress in CNN architectures and data augmentation methods, using U-Nets \cite{ronneberger_15_u-net}, FusionNets \cite{quan2016fusionnet} or inception modules \cite{beier2017multicut}. Subsequent postprocessing with the GALA algorithm \cite{GALA,knowles2016rhoananet}, conditional random fields \cite{uzunbacs_14_optree} or the lifted multicut \cite{beier2017multicut} pushed the envelope of final segmentation quality. MaskExtend \cite{meirovitch2016multi} applied CNNs to both boundary map prediction and superpixel merging, while flood-filling networks \cite{floodfill} eliminated superpixels altogether by training a recurrent neural network to perform region growing one region at a time.

Most networks mentioned so far learn boundary maps on pixels, but learning works equally well for edge-based watersheds, as was demonstrated in \cite{zlateski2015image,parag2017anisotropic} using edge weights generated with a CNN \cite{turaga2010convolutional,MALIS}. Tayloring the learning objective to the needs of the watershed algorithm by penalizing critical edges along minimax paths \cite{MALIS} or end-to-end training of edge weights and region growing \cite{wolf2017learned} improved results yet again.

Outside of connectomics, \cite{bai2016deep_watershed} obtained superior boundary maps from CNNs by learning not just boundary strength, but also its gradient direction. Holistically-nested edge detection \cite{xie2015holistically,kokkinos2015pushing} couples the CNN loss at multiple resolutions using deep supervision and is successfully used as a basis for watershed segmentation of medical images in \cite{cai2016pancreas}. 

We adopt important ideas from this prior work (hierarchical single-linkage clustering, attractive and repulsive interactions, long-range edges, and CNN-based learning). The proposed efficient segmentation framework can be interpreted as a generalization of \cite{malmberg2011generalized}, because we also allow for soft repulsive interactions (which can be overridden by strong attractive edges), and constraints are generated on-the-fly.

\section{The Mutex Watershed Algorithm as an Extension of Seeded Watershed} \label{3_methods}
In this section we introduce the Mutex Watershed Algorithm, an efficient graph clustering algorithm that can ingest both attractive and repulsive cues. We first reformulate seeded watershed as a graph partitioning with infinitely repulsive edges and then derive the generalized algorithm for finitely repulsive edges, which obviates the need for seeds.

\subsection{Definitions and notation}\label{sec:notation}
Let $\mathcal{G}=(V, E, w)$ be a weighted graph. 
The scalar attribute $w : E \rightarrow \mathbb{R}$ associated with each edge is a merge affinity: the higher this number, the higher the inclination of the two incident vertices to be assigned to the same cluster. Conversely, large negative affinity indicates a greater desire of the incident vertices to be in different clusters. In our application, each vertex corresponds to one pixel in the image to be segmented. 
We call an edge $e \in E$ repulsive if $w_{e}<0$ and we call it attractive if $w_{e}>0$ 
and collect them in $E^{-} = \{  e \in E \,|\, w_e < 0 \}$ and $E^{+} = \{ e \in E \; | \; w_e > 0 \}$ respectively.

In our application, each vertex corresponds to one pixel in the image to be segmented. 
The Mutex Watershed algorithm, defined in \autoref{3_3_MWS}, maintains disjunct active sets $A^+ \subseteq E^+$, $A^- \subseteq E^-$, $A^+ \cap A^- = \emptyset$ that encode merges and mutual exclusion constraints, respectively. Clusters are defined via the ``connected'' predicate: %
\begin{eqnarray*}
\forall i,j\in V:\\
\Pi_{i \rightarrow j} & = & \{ \text{paths } \pi\text{ from }i\text{ to }j\text{ with }\pi\subseteq E^+\}\\
\operator{connected}(i, j; A^+)&\Leftrightarrow& \exists \text{ path } \pi \in \Pi_{i \rightarrow j} \text{ with }\pi \subseteq A^+ \\
\operator{cluster}(i; A^+) & = & \{i\} \cup \{j: \operator{connected}(i,j; A^+) \}
\end{eqnarray*}
Conversely, the active subset $A^-\subseteq E^-$ of repulsive edges defines mutual exclusion relations by using the following predicate: %
\begin{eqnarray*}
\operator{mutex}(i, j; A^+, A^-)&\Leftrightarrow& \exists \, e=(k,l)\in A^- \text{ with } \\ && k \in \operator{cluster}(i; A^+) \text{ and}\\
& & l \in \operator{cluster}(j; A^+) \text{ and} \\
&&\operator{cluster}(i; A^+) \neq \operator{cluster}(j; A^+)
\end{eqnarray*}
Admissible active edge sets $A^+$ and $A^-$ must be chosen such that the resulting clustering is consistent, i.e.\ nodes engaged in a mutual exclusion constraint cannot be in the same cluster:
$\operator{mutex}(i,j; A^+, A^-) \Rightarrow \operator{not} \:\: \operator{connected}(i, j; A^+)$.
The ``connected'' and ``mutex'' predicates can be efficiently evaluated using a union find data structure.

\begin{figure}[t]
\centering
\includegraphics[width=\linewidth]{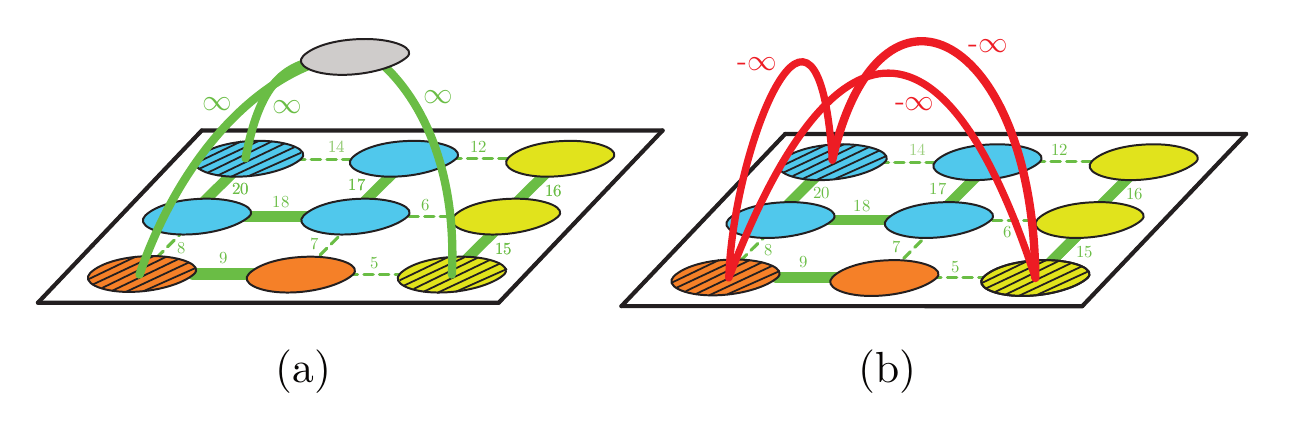}%
   \caption{Two equivalent representations of the seeded watershed clustering obtained using (a) a maximum spanning tree computation or (b) Algorithm \ref{WS_algo_code}. Both graphs share the weighted attractive~(green) edges and seeds (hatched nodes). The infinitely attractive connections to the auxiliary node (gray) in (a) are replaced by infinitely repulsive~(red) edges between each pair of seeds in (b). The two final clusterings are defined by the active sets (bold edges) and are identical. Node colors indicate the clustering result, but are arbitrary.}
\label{fig:WS_compare}
\end{figure}

\begin{algorithm}[b]
 \hrulefill \\
\textbf{Seeded Watershed:}\\
\SetKwProg{WS}{WS}{$\big(\mathcal{G}(V,E),$ \text{\rm pos.\ weights} $w:E\rightarrow \mathbb{R}^+$, \text{\rm seeds} $S\subseteq V\big)$:}{}
\WS{}{
$A^+ \leftarrow \emptyset$\; 
$A^- \leftarrow \{ (s,t) \in S \times S \,|\, s \neq t \} $\;
\Comment{\parbox[t]{.77\linewidth}{Equivalent to introducing infinitely \\repulsive edges between seeds}}\\
 \For{$(i,j) = e \in E $ {\rm in descending order of } $w_e$} 
 {
    \If{\rm {\bf not} $\operator{connected}$($i, j; A^+$) \textbf{and not} $\operator{mutex}$($i,j;A^+,A^-$) }  
    {
          $A^+ \leftarrow A^+ \cup e$ \;
          \Comment{\parbox[t]{.65\linewidth}{merge $i$ and $j$ and inherit the mutex\\ constraints of the parent clusters}}
    }
    }
    \textbf{return} $A^+ \cup A^-$
  } \vspace{5pt}\hrulefill
  \vspace{5pt}
 \caption{Mutex version of seeded watershed algorithm. The output clustering is defined by the connected components of the final attractive active set $A^+$.}
 \label{WS_algo_code}
\end{algorithm}

\subsection{Seeded watershed from a mutex perspective}
\noindent One interpretation of the proposed method is in terms of a generalization of the edge-based watershed algorithm \cite{Meyer1994,Meyer1994minimum,meyer1999morphological}
 or image foresting transform \cite{falcao2004image}.
This algorithm can only ingest a graph with purely attractive interactions, $E^{-} = \emptyset$. Without  further constraints, the algorithm would yield only the trivial result of a single cluster comprising all vertices. To obtain more interesting output, an oracle needs to provide seeds~(e.g.\ one node per cluster). These seed vertices are all connected to an auxiliary node (see Fig.~\ref{fig:WS_compare} (a)) by auxiliary edges with infinite merge affinity. A maximum spanning tree (MST) on this augmented graph can be found in linearithmic time; and the maximum spanning tree (or in the case of degeneracy: at least one of the maximum spanning trees) will include the auxiliary edges. When the auxiliary edges are deleted from the MST, a forest results, with each tree representing one cluster \cite{meyer1999morphological,Meyer1994,falcao2004image}.

We now reformulate this well-known algorithm in a way that will later emerge as a special case of the proposed Mutex Watershed: 
we eliminate the auxiliary node and edges, and replace them by a set of infinitely repulsive edges, one for each pair of seeds (Fig.~\ref{fig:WS_compare} (b)).
Algorithm \ref{WS_algo_code} is a variation of Kruskal's MST algorithm operating on the seed mutex graph just defined, and gives results identical to seeded watershed on the original graph.

This algorithm differs from Kruskal's only by the check for mutual exclusion in the if-statement. Obviously, the modified algorithm has the same effect as the original algorithm, because the final set $A^+$ is exactly the maximum spanning forest obtained after removing the auxiliary edges from the original solution. 

In the sequel, we generalize this construction by admitting less-than-infinitely repulsive edges. Importantly, these can be dense and are hence much easier to estimate automatically than seeds with their strict requirement of only-one-per-cluster.

\begin{algorithm}[t]
 \hrulefill \\
\textbf{Mutex Watershed:}\\
\SetKwProg{MWS}{MWS}{$\big(\mathcal{G}(V,E)$, $w:E\rightarrow \mathbb{R}$, \text{\rm boolean $\operator{connect\_all}\big)$}:}{}
\MWS{}{
     $A^+ \leftarrow \emptyset; \quad A^- \leftarrow \emptyset $\;
 \For{$(i,j) = e \in E $ {\rm in descending order of } $|w_e|$} 
 {
   \eIf{$e \in E^+$}
   {
   \If{\rm \textbf{not} $\operator{mutex}$($i,j; A^+, A^-$) }  
    {
        \If{\rm {\bf not} $\operator{connected}$($i, j; A^+$) \textbf{or}
            \rm $\operator{connect\_all}$}{
          $\operator{merge}(i, j)$: $A^+ \leftarrow A^+ \cup e$\;
          \Comment{\parbox[t]{.55\linewidth}{ merge $i$ and $j$ and inherit the mutex \\constraints of the parent clusters}}
        }
    }
   }{
    \If{\rm {\bf not} $\operator{connected}$($i, j; A^+$)}
      {
          $\operator{addmutex}(i, j)$: $A^- \leftarrow A^- \cup e$\;
          \Comment{\parbox[t]{.60\linewidth}{add mutex constraint between $i$ and $j$}}
      }
   }

 }
 \textbf{return} $A^+ \cup A^-$
 } \vspace{3pt}\hrulefill
 \vspace{6pt}
 \caption{Mutex Watershed Algorithm. \REVIEW{The output clustering is defined by the connected components of the final attractive active set $A^+$. The $\operator{connect\_all}$ parameter changes the internal cluster connectedness from trees to fully connected, but does not change the output clustering.} The $\operator{connected}$ predicate can be efficiently evaluated using union find data structures.}
 \label{algo_code_efficient}
\end{algorithm}

\begin{figure}[t]
\centering
\includegraphics[width=\linewidth]{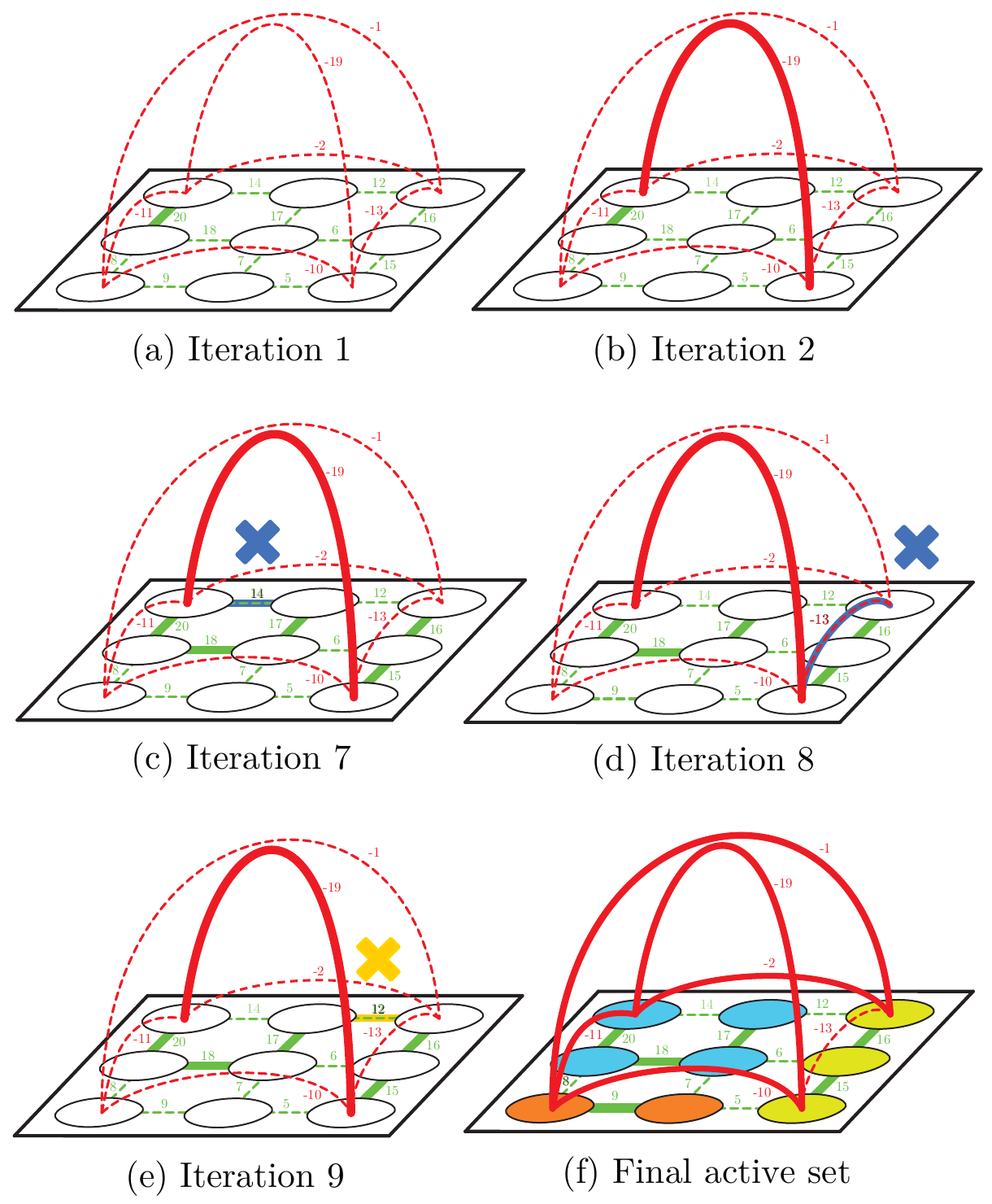}
   \caption{Some iterations of the Mutex Watershed Algorithm \ref{algo_code_efficient} applied to a graph with weighted attractive (green) and repulsive (red) edges. Edges accumulated in the active set $A$ after a given number of iterations are shown in bold. \REVIEW{The $\operator{connect\_all}$ parameter of the algorithm is set to $\operator{False}$, so that only the positive edges belonging to the maximum spanning tree of each cluster are added to the active set.} Once the algorithm terminates, the final active set (f) defines the final clustering (indicated using arbitrary node colors). Some edges are not added to the active set because they are mutex constrained~(yellow highlight) or because the associated nodes are already connected and in the same cluster~(blue highlight).}
\label{fig:MWS_algorithm}
\end{figure}

\subsection{Mutex Watersheds}\label{3_3_MWS}

\noindent 
We now introduce our core contribution: an algorithm that is empirically no more expensive than a MST computation; but that can ingest both attractive and repulsive cues and partition a graph into a number of clusters that does not need to be specified beforehand. \REVIEW{Neither seeds nor hyperparameters that implicitly determine the number of resulting clusters are required.}

The Mutex Watershed, Algorithm \ref{algo_code_efficient}, proceeds as follows. Given a graph $\mathcal{G}=(V, E)$ with signed weights $w: E \rightarrow \mathbb{R}$, do the following: sort all edges $E$, attractive or repulsive, by their absolute weight in descending order into a priority queue. Iteratively pop all edges from the queue and add them to the active set one by one, provided that a set of conditions are satisfied.
More specifically, assuming $\operator{connect\_all}$ is $\operator{False}$, if the next edge popped from the priority queue is attractive and its incident vertices are not yet in the same tree, then connect the respective trees provided this is not ruled out by a mutual exclusion constraint. If on the other hand the edge popped is repulsive, and if its incident vertices are not yet in the same tree, then add a mutual exclusion constraint between the two trees. 
The output clustering is defined by the connected components of the final attractive active set $A^+$.

 The crucial difference to Algorithm \ref{WS_algo_code} is that mutex constraints are no longer pre-defined, but created dynamically whenever a repulsive edge is found. However, new exclusion constraints can never override earlier, high-priority merge decisions. In this case, the repulsive edge in question is simply ignored. Similarly, an attractive edge must never override earlier and thus higher-priority must-not-link decisions. 

\REVIEW{The boolean value of the $\operator{connect\_all}$ input parameter of the algorithm does not influence the final output clustering, but defines the internal cluster connectedness: when it is set to $\operator{True}$, the algorithm adds all attractive intra-cluster edges to the active set $A^+$. When it is set to $\operator{False}$, then a maximum spanning tree is built for each cluster similarly to the seeded watershed. This variant of the algorithm will be helpful in the next section \ref{sec:MWS_objective} to highlight the relation between the Mutex Watershed and the multicut problem.}

Fig. \ref{fig:MWS_algorithm} illustrates the proposed algorithm: Fig. \ref{fig:MWS_algorithm}a and Fig. \ref{fig:MWS_algorithm}b show examples of an unconstrained merge and an added mutex constraint, respectively; Fig. \ref{fig:MWS_algorithm}c and Fig. \ref{fig:MWS_algorithm}d show, respectively, an example of an attractive edge ($w_e=14$) and repulsive edge ($w_e=-13$) that are not added to the active set because their incident vertices  are already ``connected'' and belong to the same tree of the forest $A^+$; finally, Fig. \ref{fig:MWS_algorithm}e shows an attractive edge ($w_e=12$) that is ruled out by a previously introduced mutual exclusion relation.

\subsection{Time Complexity Analysis}

Before analyzing the time complexity of algorithm \ref{algo_code_efficient} we first review the complexity of Kruskal's algorithm. Using a union-find data structure~(with path compression and union by rank) the time complexity of $\operator{merge}(i,~j)$ and $\operator{connected}(i,~j)$ is $\mathcal{O}(\alpha(V))$, where $\alpha$ is the slowly growing inverse Ackerman function, and the total runtime complexity is dominated by the initial sorting of the edges $\mathcal{O}(E \log E)$\cite{cormen2009introduction}.

\noindent To check for mutex constraints efficiently, we maintain a set of all active mutex edges $$ M[C_i] = \{(u,~v) \in A^- | u \in C_i \lor v \in C_i\} $$ for every $C_i = \operator{cluster}(i)$ using hash tables, where insertion of new mutex edges~(i.e. $\operator{addmutex}$) and search have an average complexity of $\mathcal{O}(1)$. Note that every cluster can be efficiently identified by its union-find root node.
For $\operator{mutex}(i,~j)$ we check if $M[C_i] \cap M[C_j] = \emptyset$ by searching for all elements of the smaller hash table in the larger hash table. Therefore $\operator{mutex}(i,~j)$ has an average complexity of $\mathcal{O}(\min (|M[C_i]|, |M[C_j]|)$. %
Similarly, during $\operator{merge}(i,~j)$, mutex constraints are inherited  by merging two hash tables, which also has an average complexity $\mathcal{O}(\min (|M[C_i]|, |M[C_j]|)$. 

\noindent In conclusion, the average runtime contribution of attractive edges $\mathcal{O}(\max(|E^{+}| \cdot\alpha(V), |E^{+}| \cdot M))$ (checking mutex constraints and possibly merging) and repulsive edges
 $\mathcal{O}(\max(|E^{-}| \cdot \alpha(V), |E^{-}|)) $ (insertion of one mutex edge) result in a total average runtime complexity of algorithm \ref{algo_code_efficient}:
 \vspace{-0.1cm}
    \begin{equation}
         \mathcal{O}(\max(E \log E \;,\; EM)).
    \end{equation}
where $M$ is the expected value of $\min (|M[C_i]|, |M[C_j]|)$ and $\alpha(V) \in \mathcal{O}(\log V) \in \mathcal{O}(\log E) $ \footnote{In the worst case $G$ is a fully connected graph, with $|E|=|V|^2$, hence $\log |V| = \frac{1}{2} \log |E|$.}.

\noindent In the worst case $\mathcal{O}(M) \in \mathcal{O}(E)$, the Mutex Watershed Algorithm has a runtime complexity of  $\mathcal{O}(E^2)$. Empirically, we find that $\mathcal{O}(EM) \approx \mathcal{O}(E\log E)$ by measuring the runtime of Mutex Watershed for different sub-volumes of the ISBI challenge~(see Figure \ref{fig:scalingM}), leading to a 
\begin{equation}
    \text{Empirical Mutex Watershed Complexity: }\mathcal{O}(E \log E)
\end{equation}

\begin{figure}
   \centering
\adjustbox{width=0.98\linewidth}{
       \input{fig/runtime/runtime.pgf}}
   \caption{Runtime $T$ of Mutex Watershed (without sorting of edges) measured on sub-volumes of the ISBI challenge of different sizes~(thereby varying the total number of edges $E$). We plot $\frac{T}{|E|}$ over $|E|$ in a logarithmic plot, which makes $T \sim |E| log(|E|)$ appear as straight line. A logarithmic function (blue line) is fitted to the measured $\frac{T}{|E|}$ (blue circles) with ($R^2 = 0.9896$). The good fit suggests that empirically $T \approx \mathcal{O}(E \log E)$.}
\label{fig:scalingM}
\end{figure}

\begin{figure*}[t]
    \centering 
    \includegraphics[width=\linewidth]{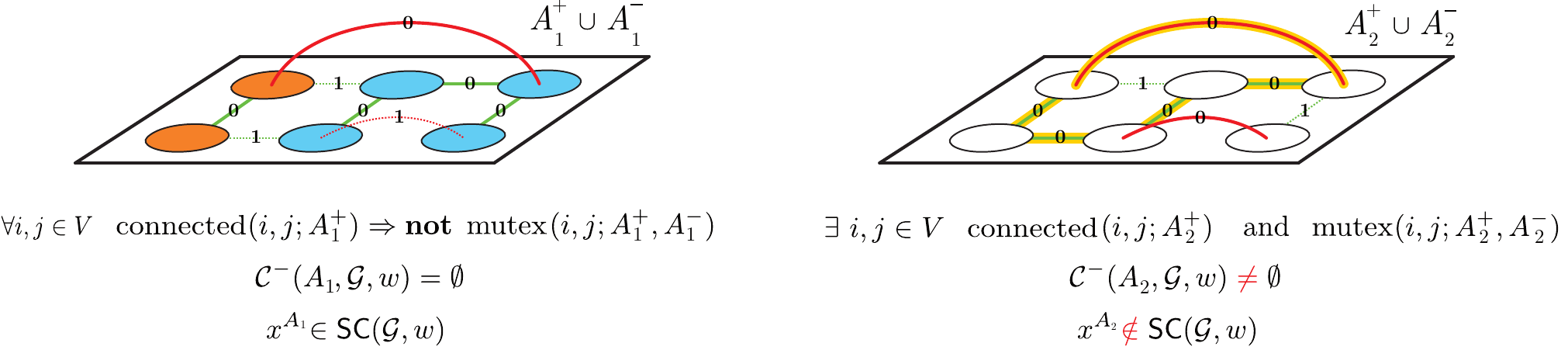}
    \caption{\textbf{Consistent and inconsistent active sets} -- Two different active edge sets $A_1\subseteq E$ (on the left) and $A_2\subseteq E$ (on the right) on identical toy graphs with six nodes, attractive~(green) and repulsive~(red) edges. The value of the edge indicator $x^A\in \{0,1\}^{|E|}$ defined in Eq.~\ref{def:active_edge_indicator} is shown for every edge. Members of the active sets are shown as solid lines.  \textbf{On the left}, the active set $A_1$ is \emph{consistent}, i.e. does not include any conflicted cycle $\mathcal{C}^-(\mathcal{G},w)$ (see Def. \ref{def:conflicted_cycles}): Therefore, it is associated with a clustering (represented by arbitrary node colors). \textbf{On the right}, the active set $A_2$ is not consistent and includes at least one conflicted cycle (highlighted in yellow), thus it cannot be associated with a node clustering.}
\label{fig:conflicted_cycles}
\end{figure*}

\section{Theoretical characterization}\label{sec:MWS_objective}

\noindent \textbf{Towards the Multicut framework.}
In section \ref{3_3_MWS}, we have introduced the Mutex Watershed (MWS) algorithm as a generalization of seeded watersheds and the Kruskal algorithm in particular. 
However, since we are considering graphs with negative edge weights, the MWS is conceptually closer to the multicut problem and related heuristics such as GAEC and GF \cite{levinkov2017comparative}.
Fortunately, due to the structure of the MWS it can be analyzed using dynamic programming. This section summarizes our second contribution, i.e. the proof that the Mutex Watershed Algorithm globally optimizes a precise objective related to the multicut.

\subsection{Review of the Multicut problem and its objective}
In the following, we will review the multicut problem not in its standard formulation but in the \textit{Cycle Covering Formulation} introduced in \cite{lange2018partial}, which is similar to the MWS formulation as it also considers the set of \textit{attractive} and \textit{repulsive} edges separately. 
Previously, in Sec.~\ref{sec:notation}, we defined a clustering by introducing the concept of an active set of edges \mbox{$A=A^+\cup A^-\subseteq E$} and the \operator{connected/}\operator{mutex} predicates. In particular, an active set describes a valid clustering if it does not include \textbf{both} a path of only attractive edges \textbf{and} a path with exactly one repulsive edge connecting any two nodes $i,j\in V$:
\begin{equation}
\operator{connected}(i, j; A^+) \quad \Longrightarrow \quad \textbf{not } \operator{mutex}(i, j; A^+,A^-).
\end{equation}
In other words, an active set is \emph{consistent} and describes a clustering if it does not contain any cycle with exactly one repulsive edge (known as conflicted cycles). 

\begin{definition} \label{def:conflicted_cycles}
\textbf{Conflicted cycles} --
We call a cycle of $\mathcal{G}$ conflicted w.r.t. $(\mathcal{G},w)$ if it contains precisely one repulsive edge $e \in E^-$, s.t. $w_e < 0$. We denote by $\mathcal{C}^-(\mathcal{G},w) \subseteq \mathcal{C}(\mathcal{G},w)$ the set of all conflicted cycles. Furthermore, given a set of edges $A \subseteq E$, we denote by $\mathcal{C}^-(A, \mathcal{G},w) \subseteq \mathcal{C}^-(\mathcal{G},w)$ the set of conflicted cycles involving only edges in $A$.
\end{definition}

\noindent 
From now on, in order to describe different clustering solutions in the framework of (integer) linear programs, we associate each active set $A$ with the following edge indicator $x^A$
\begin{equation}
         x^A := \mathbbm{1}\{e \notin A)\} \in \{0,1\}^{|E|}. \label{def:active_edge_indicator}
\end{equation}
In this way, the cycle-free property $\mathcal{C}^-(A, \mathcal{G},w)=\emptyset$ of an active set can be reformulated in terms of linear inequalities:
\begin{equation}
\forall C \in \mathcal{C}^-(\mathcal{G},w): \sum_{e\in E_C} x^A_e \geq 1 \quad \Longleftrightarrow \quad \mathcal{C}^-(A, \mathcal{G},w) = \emptyset. \label{eq:cycles_and_linear_ineq}
\end{equation}
In words, the active set cannot contain conflicted cycles; or vice versa, every conflicted cycle must contain at least one edge that is not part of the active set.
Following \cite{lange2018partial}, via this property we describe the space of all possible clustering solutions by defining the convex hull $\mathsf{SC}(\mathcal{G}, w)$ of all edge indicators corresponding to valid clusterings of $(\mathcal{G}, w)$:
\begin{definition}\label{def:set_covering_hull}
Let $\mathsf{SC}(\mathcal{G}, w)$ denote the convex hull of all edge indicators \mbox{$x \in \{0,1\}^{|E|}$} satisfying the following system of inequalities:
\begin{equation} 
\forall C \in \mathcal{C}^-(\mathcal{G},w): \quad \sum_{e\in E_C} x_e \geq 1. \label{eq:def_set_covering_hull}
\end{equation} 
That is, $\mathsf{SC}(\mathcal{G}, w)$ contains all edge labelings for which every conflicted cycle is broken at least once. We call $\mathsf{SC}(\mathcal{G}, w)$ the set covering polyhedron with respect to conflicted cycles, similarly to \cite{lange2018partial}.
\end{definition}

\noindent Fig.~\ref{fig:conflicted_cycles} summarizes these definitions and provides an example of consistent and inconsistent active sets with their associated clusterings and edge indicators.  

As shown in \cite{lange2018partial}, the \emph{multicut optimization problem} can be formulated with constraints over conflicted cycles in terms of the following integer linear program (ILP), which is NP-hard:
\begin{gather}
\min_{x \in \mathsf{SC}(\mathcal{G},w)} \sum_{e\in E} |w_e| x_e.
\label{eq:MC_set_covering_problem}
\end{gather}
The solution of the multicut problem is given by the clustering associated to the connected components of the active set $\hat{A}^+=\{e\in E^+|\hat{x}_e=0\}$, where \mbox{$\hat{x} \in \{0,1\}^{|E|}$} is the solution of (\ref{eq:MC_set_covering_problem}).

\subsection{Mutex Watershed Objective}\label{sec:MWS_objective_sub}

We now define the Mutex Watershed objective that is minimized by the Mutex Watershed Algorithm~(proof in \autoref{sec:optimality_MWS}) and show how it is closely related to the multicut problem defined in Eq. (\ref{eq:MC_set_covering_problem}). 
Lange et al.\ \cite{lange2018partial} introduce the concept of dominant edges in a graph. For example, an attractive edge $f \in E^+$ is called dominant if 
there exists a cut $B$ with $f \in E_B$ such that 
$|w_{f}| \geq \sum_{e \in E_{B} \backslash\{f\}}\left|w_{e}\right|$.
\REVIEW{These highlight an aspect of the multicut problem that can be used to search for optimal solutions more efficiently.}
Not all weighted graphs contain dominant edges; but if, assuming no ties, we raise all graph weights to a large enough power a similar property emerges.
\begin{definition}\label{def:pineq}
\textbf{Dominant power:}
Let $\mathcal{G} = (V, E, w)$ be an edge-weighted graph, with unique weights $w: E \rightarrow \mathbb{R}$. We call $p \in \mathbb{N}^+$ a dominant power if:
\begin{equation}
    |w_e|^p > \sum_{t \in E,\; w_t < w_e} |w_t|^p \qquad \forall e \in E, \label{eq:pcondition}
\end{equation}
\end{definition}
 \noindent In contrast to dominant edges \cite{lange2018partial}, we do not consider edges on a cut but rather all edges with smaller absolute weight. Note that there exists a dominant power for any finite set of edges, since for any $e \in E$ we can divide (\ref{eq:pcondition}) by $|w_e|^p$ and observe that the normalized weights $|w_t|^p/|w_e|^p$ (and any finite sum of these weights) converges to 0 when $p$ tends to infinity. 

 By considering the multicut problem in Eq. (\ref{eq:MC_set_covering_problem}) and raising the weights $|w_e|$ to a dominant power $p$, we fundamentally change the problem structure:
 \begin{definition}\label{def:MWS_objective}
\textbf{Mutex Watershed Objective:}
Let $\mathcal{G} = (V, E, w)$ be an edge-weighted graph, with unique weights $w:E \rightarrow \mathbb{R}$ and $p \in \mathbb{N}^+$ a dominant power. Then the Mutex Watershed Objective is defined as the integer linear program
\begin{equation}
  \min_{x \in \mathsf{SC}(\mathcal{G}, w)} \quad \sum_{e \in E}  |w_e|^p \, x_e  \label{eq:MWS_objective_SC}
\end{equation}
where $\mathsf{SC}(\mathcal{G}, w)$ is the convex hull defined in Def.~\ref{def:set_covering_hull}.
\end{definition}

In the following section, we will prove that this modified version of the multicut objective, which we call Mutex Watershed Objective, is indeed optimized by the Mutex Watershed Algorithm:

\begin{restatable}{theorem}{Objective}\label{theo:optimal_v1}
Let $\mathcal{G} = (V, E, w)$ be an edge-weighted graph, with unique weights $w:E \rightarrow \mathbb{R}$ and $p \in \mathbb{N}^+$ a dominant power. Then the edge indicator given by the Mutex Watershed Algorithm \ref{algo_code_efficient} $$x^{\mathbf{MWS}} := \mathbbm{1}\Big\{e \notin \mathbf{MWS}\Big(\mathcal{G}, w, \operator{connect\_all=True}\Big)\Big\}$$ minimizes the Mutex Watershed Objective in Eq.~(\ref{eq:MWS_objective_SC}).
\end{restatable} 
\noindent 

\subsection{Proof of optimality via dynamic programming} \label{sec:optimality_MWS}

\begin{algorithm}[b]
 \hrulefill \\
 \textbf{Conflicted-Cycles Mutex Watershed:}\\
\SetKwProg{CCMWS}{CCMWS}{$\big(\mathcal{G}(V,E),w:E\rightarrow \mathbb{R}\big)$:}{}
\CCMWS{}{
 $A \leftarrow \emptyset$\;
 \For{$(i,j) = e \in E $ {\rm in descending order of } $|w_e|$} 
 {
    \If{\rm $\mathcal{C}^-(A\cup \{e\},\mathcal{G},w) = \emptyset$}{
          $A \leftarrow A \cup e$\;
    
    }
 }
 \Return $A$\;
 }
  \vspace{3pt}
\hrulefill
 \vspace{3pt}
 \caption{Equivalent formulation of the Mutex Watershed Algorithm \ref{algo_code_efficient}, \REVIEW{with input parameter $\operator{connect\_all=True}$}. The set of conflicted cycles $\mathcal{C}^-(A,\mathcal{G},w)$ is defined in Def. \ref{def:conflicted_cycles}. The output clustering is defined by the connected components of the final attractive active set $A^+=A\cap E^+$.}
 \label{MWS_conflicted_cycles}
\end{algorithm}

In this section we prove Theorem \ref{theo:optimal_v1}, i.e.\ that the Mutex Watershed Objective defined in \ref{def:MWS_objective} is solved to optimality by the Mutex Watershed Algorithm \ref{MWS_conflicted_cycles}. Particularly, in the following Sec.~\ref{sec:cycle_consistency} we show that the edge indicator associated to the solution of the MWS algorithm lies in $\mathsf{SC}(\mathcal{G},w)$, whereas in Sec.~\ref{sec:optimilaty_proof} we prove that it solves Eq.~\ref{eq:MWS_objective_SC} to optimality.

\subsubsection{Cycle consistency}\label{sec:cycle_consistency}
The Mutex Watershed algorithm introduced in Sec.~\ref{3_methods} iteratively builds an active set $A = A^+ \cup A^-$ such that nodes engaged in a mutual exclusion constraint (encoded by edges in $A^-$) are never part of the same cluster. In other words, this means that the active set built by the Mutex Watershed at every iteration does never include a \emph{conflicted cycle} and is always \emph{consistent}. 
In particular, for any attractive edge $(i,j) = e^+ \in E^+$ and any consistent set $A$ that fulfills $\mathcal{C}^-(A,\mathcal{G},w) = \emptyset$:
\begin{align*}
\mathbf{not }\operator{ mutex}(i, j, A^+, A^-)\quad \Leftrightarrow &\quad \mathcal{C}^-(A\cup \{e^+\},\mathcal{G},w) = \emptyset 
\intertext{Similarly, for any repulsive edge $(s,t) = e^- \in E^-$:}
\mathbf{not} \operator{ connected}(s, t, A^+) \quad \Leftrightarrow &\quad \mathcal{C}^-(A\cup \{e^-\},\mathcal{G},w) = \emptyset
\end{align*}

\noindent Therefore, we can rewrite Algorithm~\ref{algo_code_efficient}  in the form of Algorithm~\ref{MWS_conflicted_cycles}. 
This new formulation makes it clear that 
\begin{equation}
    \mathcal{C}^-\Big(\mathbf{MWS}\big(\mathcal{G}, w, \operator{connect\_all=True}\big)\Big) = \emptyset.
\end{equation}
Thus, thanks to Eq.~\ref{eq:cycles_and_linear_ineq} and definition \ref{def:set_covering_hull}, it follows that the MWS edge indicator $x^{\mathbf{MWS}}$ defined in \ref{theo:optimal_v1} lies in  $\mathsf{SC}(\mathcal{G},w)$:
\begin{equation}
    x^{\mathbf{MWS}} \in \mathsf{SC}(\mathcal{G},w).
\end{equation}

\subsubsection{Optimality}\label{sec:optimilaty_proof}
We first note that the Mutex Watershed Objective \ref{def:MWS_objective} and Theorem~\ref{theo:optimal_v1} can easily be reformulated in terms of active sets to minimize 
\begin{equation}
  \argmin_{A \subseteq E} \quad - \sum_{e \in A}  |w_e|^p  \qquad \operator{s.t.} \quad \mathcal{C}^-(A,\mathcal{G},w)=\emptyset.  \label{eq:full_problem_active_set}
\end{equation}

\noindent We now generalize the Mutex Watershed (see Algorithm~\ref{algo_subproblems}) and the objective such that an initial consistent set of active edges $\Ainit \subseteq E$ is supplied:

\begin{definition} \label{def:general_MWS_obj}
\textbf{Energy optimization subproblem.}
Let \mbox{$\mathcal{G} = (V, E, w)$} be an edge-weighted graph. Define the optimal solution of the subproblem as
\begin{gather} \label{eq:general_MWS_obj}
  \subproblem(\mathcal{G},\Ainit) := \underset{A \subseteq (E \setminus \Ainit)}{\text{argmin}} \; T(A) \qquad \operator{with} \quad T(A) := - \sum_{e \in A} |w_e|^p, \\
  \operator{s.t.} \quad \mathcal{C}^{-}(A \cup \Ainit, \mathcal{G}, w) = \emptyset, \label{eq:def_energy}
\end{gather}
 where $\Ainit \subseteq E$ is a set of initially activated edges such that $\mathcal{C}^{-}(\Ainit, \mathcal{G}, w) = \emptyset$. 
\end{definition}
\noindent We note that for $\Ainit = \emptyset$, the optimal solution $\subproblem(\mathcal{G},\emptyset)$ is equivalent to the solution minimizing the Mutex Watershed Objective and Eq. (\ref{eq:full_problem_active_set}).

\begin{algorithm}[t]
 \hrulefill \\
  \textbf{Initialized Mutex Watershed:}\\
  \SetKwProg{IMWS}{IMWS}{$\big(\mathcal{G}(V,E), w:E\rightarrow \mathbb{R}$, {\color{blue}\rm initial active set $\Ainit $}$\big)$:}{}
  \IMWS{}{
     $A \leftarrow \emptyset$\;
 \For{$ {\color{blue}e \in E \setminus \Ainit} $ {\rm in descending order of weight}} 
 {
    \If{\rm $\mathcal{C}^{-}({\color{blue}A\cup \Ainit \cup \{e\}}, \mathcal{G}, w) = \emptyset$}{
          $A \leftarrow A \cup e$\;
    
    }
 }
 \Return $A$\;
 }
  \vspace{3pt}
\hrulefill
 \vspace{3pt}
 \caption{Mutex Watershed algorithm starting from initial active set $\Ainit$. An initial set $\Ainit$ of active edges is given as additional input and the final active set is such that $A \subseteq E \setminus \Ainit$. Note that Algorithm \ref{MWS_conflicted_cycles} is a special case of this algorithm when $\Ainit= \emptyset$. Differences with Algorithm \ref{MWS_conflicted_cycles} are highlighted in blue.}
 \label{algo_subproblems}
\end{algorithm}

\begin{definition}
\textbf{Incomplete, consistent initial set:}\label{def:initialset}
For an edge-weighted graph $\mathcal{G} = (V, E, w)$  a set of edges $\Ainit \subseteq E$ is consistent if
\begin{equation}
    \mathcal{C}^{-}(\Ainit, \mathcal{G}, w) = \emptyset.
\end{equation}
$\Ainit$ is incomplete if it is not the final solution and there exists a consistent edge $\tilde{e}$ that can be added to $\Ainit$ without violating the constraints.
    \begin{equation}
 \exists\, \tilde{e} \in E \setminus \Ainit \quad \operator{s.t.} \quad  \mathcal{C}^{-}(\Ainit \cup \{\tilde{e}\}, \mathcal{G}, w) = \emptyset
\end{equation}
\end{definition}

\begin{definition}
\textbf{First greedy step:}\label{def:greedy_step}
Let us consider an incomplete, consistent initial active set $\Ainit \subseteq E$ on $\mathcal{G} = (V, E, w)$. We define 
\begin{equation}\label{eq:greedy_step}
       g := \underset{e \in (E \setminus \Ainit)}{\text{argmax}} \;  \; |w(e)| \quad \operator{ s.t. } \quad \mathcal{C}^{-}(\Ainit \cup \{e\}, \mathcal{G}, w)  =  \emptyset. 
\end{equation}
as the feasible edge with the highest weight, which is always the first greedy step of Algorithm \ref{algo_subproblems}.
\end{definition}
\noindent In the following two lemmas, %
 we prove that the Mutex Watershed problem has an \emph{optimal substructure property} and a \emph{greedy choice property} \cite{cormen2009introduction},
which are sufficient to prove that the Mutex Watershed algorithm finds the optimum of the Mutex Watershed Objective.

\begin{lemma} \label{theo:greedy_choice}
\textbf{Greedy-choice property.}
For an incomplete, consistent initial active set $\Ainit$ of the Mutex Watershed, the first greedy step $g$ is always part of the optimal solution $$ g \in \subproblem(\mathcal{G},\Ainit).$$%
\end{lemma}
\begin{proof}

\noindent We will prove the theorem by contradiction by assuming that the first greedy choice is not part of the optimal solution, i.e. $g \notin \subproblem(\mathcal{G},\Ainit)$. 
Since $g$ is by definition the feasible edge with highest weight, it follows that:
\begin{equation}
|w(e)| < |w(g)| \quad \forall e \in \subproblem(\mathcal{G},\Ainit).%
\end{equation}
\noindent We now consider the alternative active set $A' = \{g\}$, that is a consistent solution, with 
\begin{equation}
T(A') = -|w_g|^p \overset{(\ref{eq:pcondition})}{<} -\sum_{t \in \subproblem(\mathcal{G},\Ainit)} |w_t|^p = T\Big(\subproblem(\mathcal{G},\Ainit)\Big)
\end{equation}
which contradicts the optimality of $\subproblem(\mathcal{G},\Ainit)$.
\end{proof}

\begin{lemma}\label{theo:optm_sub_prop}
\textbf{Optimal substructure property.}
Let us consider an initial active set $\Ainit$, the optimization problem defined in Equation \ref{eq:general_MWS_obj}, and assume to have an incomplete, consistent problem (see Def. \ref{def:initialset}). Then it follows that:
\begin{enumerate}
\item After making the first greedy choice $g$, we are left with a subproblem that can be seen as a new optimization problem of the same structure;
\item The optimal solution $\subproblem(\mathcal{G},\Ainit)$ is always given by the combination of the first greedy choice and the optimal solution of the remaining subproblem.
\end{enumerate}
 
\end{lemma}

\begin{proof}
After making the first greedy choice and selecting the first feasible edge $g$ defined in Equation \ref{eq:greedy_step}, we are clearly left with a new optimization problem of the same structure that has the following optimal solution: $\subproblem(\mathcal{G},\Ainit \cup \{g\})$. \\
In order to prove the second point of the theorem, we now show that:
\begin{equation}\label{eq:optimal_sub}
\subproblem(\mathcal{G},\Ainit) = \{g\} \; \cup \subproblem(\mathcal{G},\Ainit \cup \{g\}).
\end{equation}
Since algorithm \ref{algo_subproblems} fulfills the greedy-choice property, ${g\in \subproblem(\mathcal{G},\Ainit)}$ and we can add the edge $g$ as an additional constraint to the optimal solution:
\begin{align}
\begin{split}
  \subproblem(\mathcal{G},\Ainit) = & \underset{A \subseteq (E \setminus \Ainit)}{\text{argmin}} \quad T(A) \\
  \text{s. t.}& \quad \mathcal{C}^{-}(A \cup \Ainit, \mathcal{G}, w) = \emptyset; \quad g \in A \label{eq:optimal subproblem}
\end{split}
\intertext{Then it follows that:}
\begin{split}
  \subproblem(\mathcal{G},\Ainit) = &\; \{g\} \; \cup \underset{A \subseteq \; E \setminus (\Ainit \cup \{g\})}{\text{argmin}} \quad T(A) \\
  \text{s. t.}& \quad  \mathcal{C}^{-}\Big(A \cup \{g\} \cup \Ainit, \mathcal{G}, w \Big) = \emptyset
\end{split}
\end{align}
which is equivalent to Equation \ref{eq:optimal_sub}.
\end{proof}
\begin{proof}[\textbf{Proof of Theorems \ref{theo:optimal_v1}}]%
In Lemmas \ref{theo:greedy_choice} and \ref{theo:optm_sub_prop} we have proven that the optimization problem defined in \ref{eq:full_problem_active_set} has the optimal substructure and a greedy choice property. 
It follows through induction that the final active set $\mathbf{MWS}\big(\mathcal{G}, w, \operator{connect\_all=True}\big)$ found by the Mutex Watershed Algorithm \ref{MWS_conflicted_cycles} is the optimal solution for the Mutex Watershed objective (\ref{eq:full_problem_active_set}) \cite{cormen2009introduction}.
\end{proof}

\subsection{Relation to the extended Power Watershed framework}\label{sec:power_ws}
The Power Watershed \cite{powerws} is an important framework for graph-based image segmentation that includes several algorithms like seeded watershed, random walker and graph cuts. Recently, \cite{najman2017extending} extended the framework to even more general types of hierarchical optimization algorithms thanks to the use of $\Gamma$-theory and \mbox{$\Gamma$-convergence} \cite{dal2012introduction,braides2006handbook}.
In this section, we show how the Mutex Watershed algorithm can also be included in this extended framework\footnote{The connection between the Mutex Watershed and the extended Power Watershed framework was kindly pointed out by an anonymous reviewer.} and how the framework suggests an optimization problem that is solved by the Mutex Watershed.

\subsubsection{Mutex Watershed as hierarchical optimization algorithm}
We first start by introducing the extended Power Watershed framework and restating the main theorem from \cite{najman2017extending}: 
\begin{theorem} \label{theorem:PW_framework}
\textbf{\cite{najman2017extending} Extended Power Watershed Framework.}
Consider three strictly positive integers $p,m,t\in \mathbb{N}^+$ and $t$ real numbers 
\begin{equation}
    1 \geq \lambda_0 > \lambda_1 > \ldots \lambda_{t-1}>0 \label{eq:sorted lambda}
\end{equation}
Given $t$ continuous functions $Q_k: \mathbb{R}^m \rightarrow \mathbb{R}$ with $0\leq k < t$, define the function
\begin{align}
Q^p(x) := \sum_{0\leq k< t} \lambda^p_k Q_k (x). \label{eq:def_Q_p}
\end{align}
Then, if any sequence $(x_p)_{p>0}$ of minimizers $x_p$ of $Q^p(x)$ is bounded (i.e. there exists $C>0$ such that for all $p>0$, $||x_p||_{\infty}\leq C$), the sequence is convergent, up to taking a subsequence, toward a point of $M_{t-1}$, which is the set of minimizers recursively defined in Algorithm \ref{PWS_general_alg}.
\end{theorem}
\begin{proof}
See \cite{najman2017extending} (Theorem 3.3).
\end{proof}

\begin{algorithm}[t]
 \hrulefill \\
  \textbf{Generic hierarchical optimization:}\\
  \SetKwProg{GHO}{GHO}{($Q_0, \hdots, Q_{t-1}$):}{}
  \GHO{}{

 $M_0 = \argmin_{x \in \mathbb{R}^m} Q_0(x)$ \\
 \For{$k \in 1, \ldots,t-1 $} 
 {
  $M_k = \argmin_{x \in M_{k-1}} Q_k(x)$
 }
 }
 \textbf{Return: } some $x^* \in M_{t-1}$\\
 \vspace{-6pt}\hrulefill
 \vspace{6pt}
 \caption{Generic hierarchical optimization algorithm introduced in \cite{najman2017extending}. The sequence of continuous functions \mbox{$Q_k: \mathbb{R}^m \rightarrow \mathbb{R}$} is sorted according to the associated scales $\lambda_k$ (Eq.~\ref{eq:sorted lambda}).}
 \label{PWS_general_alg}
\end{algorithm}

We now show that the Mutex Watershed algorithm can be seen as a special case of the generic hierarchical Algorithm \ref{PWS_general_alg}, for a specific choice of scales $\lambda_k$ and functions ${Q_k(x): \mathbb{R}^{m} \rightarrow \mathbb{R}}$ (see definitions (\ref{def:scales_MWS}, \ref{eq:def_Q_k_MWS}) below) . \\

\noindent \textbf{Scales $\lambda_k$:} Let $\tilde{w}_k$ be the signed edge weights $w: E \rightarrow \mathbb{R}$ ordered by decreasing absolute value $|\tilde{w}_1| > |\tilde{w}_2| > \ldots > |\tilde{w}_{t-1}|$. If two edges share the same weight, then the weight is called $\tilde{w}_k$ for both and $E_k \subseteq E$ denotes the set of all edges with weight $\tilde{w}_k$. We then define \textit{the scales} $\lambda_k$ as 
\begin{equation}
\lambda_k := 
\begin{cases}
1 & \text{if} \,\,\, k=0 \\
\left| \frac{\tilde{w}_k}{2\tilde{w}_1} \right| & \text{otherwise.}
\end{cases}\label{def:scales_MWS}
\end{equation}

\noindent \textbf{The continuous functions $Q_k(x): \mathbb{R}^{|E|} \rightarrow \mathbb{R}$} are defined as follows
\begin{equation}
Q_k(x) := 
\begin{cases}
|E|\cdot \min_{x' \in \mathsf{\intSC}(\mathcal{G},w)} ||x'-x|| \quad & \text{if } k=0\\
\sum_{e\in E_k} x_e & \text{otherwise,}\\
\end{cases} \label{eq:def_Q_k_MWS}
\end{equation}
where $\mathsf{\intSC}(\mathcal{G},w)$ is defined as:
\begin{equation}
\mathsf{\intSC}(\mathcal{G},w) := \mathsf{SC}(\mathcal{G},w) \cap \{0,1\}^{|E|}. \label{def:\intSC}
\end{equation}
In words, $Q_0(x)$ is proportional to the distance between $x$ and the closest point on the set $\mathsf{\intSC}(\mathcal{G},w)$, whereas $Q_k(x)$ depends only on the indicators $x_e$ of edges in $E_k$, for $k>0$.

Algorithm \ref{PWS_general_alg_MWS} is obtained by substituting the scales $\lambda_k$ and functions $Q_k(x)$ (respectively defined in Eq. (\ref{def:scales_MWS}) and (\ref{eq:def_Q_k_MWS})) into Algorithm \ref{PWS_general_alg} .  
The algorithm starts by setting $M_0$ to $\mathsf{\intSC}(\mathcal{G},w)$, i.e. by restricting the space of the solutions only to integer edge labelings $x$ that do not include any conflicted cycles. Then, in the following iterations $k \in 1, \ldots,t-1 $, the algorithm solves a series of minimization sub-problems that in the most general case are NP-hard, even though they involve a smaller set of edges $E_k\subseteq E$. 
Nevertheless, if we assume that all weights are distinct, then $|E_k|=1$ for all $k$ and the solution to the sub-problems amounts to checking if the new edge can be labeled with $x_e=0$ without introducing any conflicted cycles. This procedure is identical to Algorithm \ref{algo_code_efficient}: at every iteration, the Mutex Watershed tries to add an edge to the active set $A$, provided that no mutual exclusion constraints are violated. 

In summary, the framework in \cite{najman2017extending} provides a new formulation of the Mutex Watershed Algorithm that is even applicable to graphs with tied edge weights. In practice, when edge weights are estimated by a CNN, we do not expect tied edge weights.

\begin{algorithm}[t]
\SetAlgoLined
 \hrulefill \\
  \SetKwProg{PWSMWS}{PWSMWS}{($Q_0, \hdots, Q_{t-1}$):}{}
  \PWSMWS{}{

$M_{0} = \argmin_{x \in \mathbb{R}^{|E|}} Q_0(x)  = \mathsf{\intSC}(\mathcal{G},w)$ \vspace{5pt}\\
 \For{$k \in 1, \ldots,t-1 $} 
 {
  $M_k = \argmin_{x \in M_{k-1}} \sum_{e\in E_{k}} x_e$

 }
 }\vspace{5pt}
 \textbf{Return: } some $x^* \in M_{t-1}$\\
 \vspace{-6pt}\hrulefill
 \vspace{6pt}
 \caption{Special case of the general hierarchical Algorithm \ref{PWS_general_alg} obtained by substituting Def.~(\ref{def:scales_MWS}) and (\ref{eq:def_Q_k_MWS}). With the additional assumption of unique signed weights $w:E\rightarrow \mathbb{R}$, this algorithm is equivalent to the Mutex Watershed Algorithm \ref{MWS_conflicted_cycles}. The sequence of functions \mbox{$Q_k: \mathbb{R}^m \rightarrow \mathbb{R}$} defined in Eq.~\ref{eq:def_Q_k_MWS} is sorted according to the associated scales $\lambda_k$ in Eq.~\ref{def:scales_MWS}. $\mathsf{\intSC}(\mathcal{G},w)$ is defined in Eq.~\ref{def:\intSC}}
 \label{PWS_general_alg_MWS}
\end{algorithm}

\subsubsection{Convergence of the sequence of minimizers}
In this section, we see how Theorem \ref{theorem:PW_framework} also suggests a minimization problem that is solved by the Mutex Watershed algorithm. A short summary is given in the final paragraph of the section.

\noindent First, we make sure that the conditions of Theorem \ref{theorem:PW_framework} are satisfied when we apply it to Algorithm \ref{PWS_general_alg_MWS}:
\begin{restatable}{lemma}{boundedsequence}
\label{lemma:bounded_sequence}
Let us consider the scales $\lambda_k$ and continuous functions ${Q_k(x): \mathbb{R}^{|E|} \rightarrow \mathbb{R}}$ respectively defined in Eq. (\ref{def:scales_MWS}) and (\ref{eq:def_Q_k_MWS}). For any value of $p\in \mathbb{N}^+$, let $x_p \in \mathbb{R}^{|E|}$ be a minimizer of the function $Q^p(x)$ defined in Eq. (\ref{eq:def_Q_p}). Then, the minimizer $x_p$ lies in the set $\mathsf{\intSC}(\mathcal{G},w)$. From this, it follows that any sequence of minimizers $(x_p)_{p>0}$ is bounded and the conditions of Theorem \ref{theorem:PW_framework} are satisfied.
\end{restatable}
\begin{proof}
See Appendix \ref{sec:proof_minizers}.
\end{proof}
\noindent Then, given any $p\in \mathbb{N}^+$ and the Def.~(\ref{def:scales_MWS}, \ref{eq:def_Q_k_MWS}), we have that the minimization of the function $Q^p(x)$ defined in Eq.~(\ref{eq:def_Q_p}) is given by the following problem:
\begin{align}
\argmin_{x\in \mathbb{R}^m} \, Q^p(x) = & \argmin_{x\in \mathbb{R}^m} \, \sum_{0\leq k< t} \lambda^p_k Q_k (x) \label{eq:MWS_limit_1} \\
= & \argmin_{x\in \mathsf{\intSC}(\mathcal{G},w)} \, \sum_{1 \leq k < t} \left | \frac{\tilde{w}_k}{2\tilde{w}_1} \right |^p \sum_{e\in E_{k}} x_e  \\
 = & \argmin_{x\in \mathsf{\intSC}(\mathcal{G},w)} \frac{1}{|2\tilde{w}_1|^p} \sum_{e \in E} |w_e|^p \,\,x_e \label{eq:MWS_limit_2}
\end{align}
where we used Lemma \ref{lemma:bounded_sequence} and restricted the domain of the $\argmin$ operation to $\mathsf{\intSC}(\mathcal{G},w)$, so that $Q_0(x)=0$ for all $x\in \mathsf{\intSC}(\mathcal{G},w)$.

It follows from Lemma \ref{lemma:bounded_sequence} and Theorem \ref{theorem:PW_framework} that a sequence of minimizers $(x_p)_{p>0}$ of the problem (\ref{eq:MWS_limit_2}) converge, up to taking a subsequence, to the solution $x^*$ returned by Algorithm \ref{PWS_general_alg_MWS}. 
More specifically, we know that any minimizer $x_p$ of (\ref{eq:MWS_limit_2}) is in the discrete set $\mathsf{\intSC}(\mathcal{G},w)$. Hence, the convergent sequence of minimizers $(x_p)_{p>0}$ eventually becomes constant and there exists a $p'\in \mathbb{N}^+$ large enough such that $x_p=x^*$ for all $p\geq p'$. In other words, in the case of unique weights and $p\geq p'$ large enough, the solution $x^*$ of the Mutex Watershed Algorithm \ref{PWS_general_alg_MWS} solves the problem (\ref{eq:MWS_limit_2}), which is just a rescaled version of the Mutex Watershed Objective we introduced in Sec.~\ref{sec:MWS_objective_sub}.

 To summarize, we used the extended Power Watershed framework to show that the Mutex Watershed provides a solution to the minimization problem in Eq. (\ref{eq:MWS_limit_2}) for $p$ large enough. In particular, this problem suggested by the Power Watershed framework is the same one previously derived in Sec. \ref{sec:MWS_objective_sub} by linking the Mutex Watershed Algorithm to the multicut optimization problem.

\section{Experiments} \label{4_results}

We evaluate the Mutex Watershed on the challenging task of 
neuron segmentation in electron microscopy (EM) image volumes.
This application is of key interest in connectomics, a field of 
neuro-science that strives to reconstruct neural wiring digrams spanning complete central nervous systems. The task requires segmentation of neurons from electron microscopy images of neural tissue -- a challenging endeavor, since segmentation has to be based only on boundary information (cell membranes) and some of the boundaries are not very pronounced. Besides, cells contain membrane-bound organelles, which have to be suppressed in the segmentation. Some of the neuron protrusions are very thin, but all of those need to be preserved in the segmentation to arrive at the correct connectivity graph. While a lot of progress is being made, currently only manual tracing or proof-reading yields sufficient accuracy for correct circuit reconstruction \cite{schlegel2017Learning}.

We validate the Mutex Watershed algorithm on the most popular neural segmentation challenge: ISBI2012 \cite{isbi2012challenge}. We estimate the edge weights using a CNN as described in Section \ref{4_cnn} and compare with other entries in the leaderboard as well as with other popular post-processing methods for the same network predictions in Section \ref{4_isbi}.

\subsection{Estimating edge weights with a CNN} \label{4_cnn}
The common first step to EM segmentation is to predict which pixels belong to a cell membrane using a CNN. Different post-processing methods are then used to obtain a segmentation, see \autoref{2_rel_work} for an overview of such methods.
The CNN can either be trained to predict boundary pixels \cite{ciresan_12_deep-em-segmentation,beier2017multicut} or undirected affinities \cite{lee2017superhuman,funke2018large} which express how likely it is for a pixel to belong to a different cell than its neighbors in the 6-neighborhood. In this case, the output of the network contains three channels, corresponding to left, down and next imaging plane neighbors in 3D. The affinities do not  have to be limited to immediate neighbors -- in fact, \cite{lee2017superhuman} have shown that introduction of long-range affinities is beneficial for the final segmentation even if they are only used to train the network. Building on the work of \cite{lee2017superhuman}, we train a CNN to predict short- and long-range affinities and then use those directly as weights for the Mutex Watershed algorithm. 

\begin{figure}
    \centering
    \begin{subfigure}[b]{0.95 \linewidth}
    \centering
    \includegraphics[width=0.9\linewidth,valign=t]{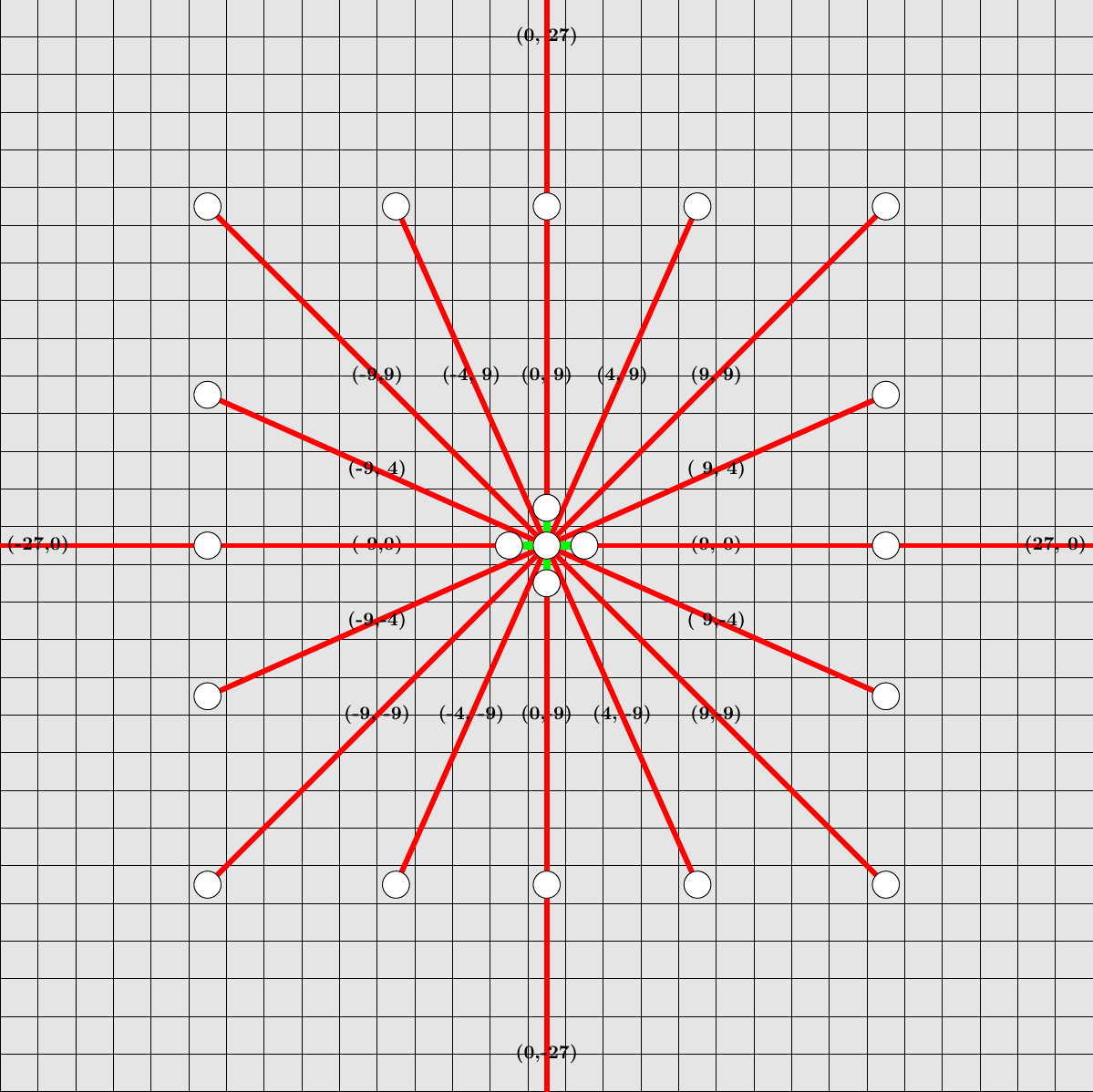}%
    \caption{XY-plane neighborhood with local attractive edges (green) and
     sparse repulsive edges (red) with approximate radius 9 and further long-range connections with distance 27}\label{fig:2d-connection}
    \end{subfigure}
    \par\bigskip
    \begin{subfigure}[b]{0.95 \linewidth}
    \centering
    \includegraphics[width=0.4\linewidth, valign=t]{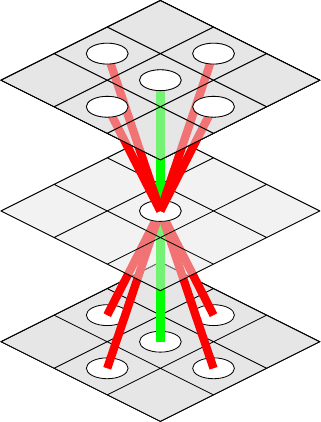}
    \caption{Due to the great anisotropy of the data we limit the Z-plane edges to a distance of 1. The direct neighbors are attractive, whereas the indirect neighbors are repulsive.}\label{fig:3d-connection}
    \end{subfigure}%
    \caption{Local neighborhood structure of attractive~(green) and repulsive~(red) edges in the Mutex Watershed graph.}
    \label{fig:connectivity}
\end{figure}

We estimate the affinities / edge weights for the neighborhood structure shown in Figure \ref{fig:connectivity}. To that end, we define local attractive and long-range repulsive edges. When attractive edges are only short-range, the solution will consist of spatially connected segments that cannot comprise ``air bridges''. This holds true for both (lifted)~multicut and for Mutex Watershed. We use a different pattern for in-plane and between-plane edges due to the great anisotropy of the data set. In more detail, we pick a sparse ring of in-plane repulsive edges and additional longer-range in-plane edges which are necessary to split regions reliably (see Figure \ref{fig:2d-connection}).
We also added connections to the indirect neighbors in the lower adjacent slice to ensure correct 3D connectivity (see Figure \ref{fig:3d-connection}). In our experiments, we pick a subset of repulsive edges, by using strides of 2 in the XY-plane in order to avoid artifacts caused by occasional very thick membranes. Note that the stride is not applied to local (attractive) edges, but only to long-range (repulsive) edges. The particular pattern used was selected after inspecting the size of typical regions. The specific pattern is the only one we have tried and was {\it not} optimized over.

In total, $C^+$ attractive and $C^-$ repulsive edges are defined for each pixel, resulting in $C^+ + C^-$ output channels in the network. 
We partition the set of attractive / repulsive edges into subsets $H^+$ and $H^-$ that contain all edges at a specific offset:
$\label{edgesets}
    E^+ = {\bigcup_{c=1}^{C^+}} H^+_{c}$ for attractive edges, with $H^{-}$ defined analogously. 
Each element of the subsets $H^+_{c}$ and $H^-_{c}$ corresponds to a specific channel predicted by the network. We further assume that weights take values in $[0,1]$.%

\subsubsection*{Network architecture and training}
We use the 3D U-Net \cite{ronneberger_15_u-net, cciccek20163d} architecture, as proposed in \cite{funke2018large}. %

Our training targets for attractive / repulsive edges $\gt{w}^\pm$ can be derived from a groundtruth label image $\gt{L}$ according to
\begin{equation}
    \gt{w}^+_{e=(i, j)} = \begin{cases}
        1 , &\text{ if } \gt{L}_i = \gt{L}_j\\
        0 , & \text{otherwise}
    \end{cases}\quad
\end{equation}
\begin{equation}
    \gt{w}^-_{e=(i, j)} = \begin{cases}
        0 , &\text{ if } \gt{L}_i = \gt{L}_j\\
        1 , & \text{otherwise}
    \end{cases}
\end{equation}

Here, $i$ and $j$ are the indices of vertices / image pixels.
Next, we define the loss terms

\begin{equation} \label{dice_plus}
    \mathcal{J}^+_{c} = - \frac{\sum_{e \in H^+_{c}} (1 - w^+_e) (1 - \gt{w}^+_e)}{\sum_{e \in H^+_{c}} ((1 - w^+_e)^2 + (1 -\gt{w}^+_e)^2)} 
\end{equation}\label{dice_minus}
\begin{equation}
 \mathcal{J}^-_{c} = - \frac{\sum_{e \in H^-_{c}} w^-_e {\gt{w}^-_e}}{\sum_{e \in H^-_{c}} ((w^-_e)^2 + (\gt{w}^-_e)^2)}
\end{equation}
for attractive edges (i.e. channels) and repulsive edges (i.e. channels).

Equation \ref{dice_plus} is the S{\o}rensen-Dice coefficient \cite{dice1945measures,sorensen1948method} formulated for fuzzy set membership values.
During training we minimize the sum of attractive and repulsive loss terms $\mathcal{J} = \sum_{c}^{C^+} \mathcal{J}^{+}_{c} + \sum_{c}^{C^-} \mathcal{J}^{-}_{c}$. This corresponds to summing up the channel-wise S{\o}rensen-Dice loss. 
The terms of this loss are robust against prediction and / or target sparsity, a desirable quality for neuron segmentation: since membranes are locally two-dimensional and thin, they occupy very few pixels in three-dimensional the volume. 
More precisely, if $w^{+}_e$ or $\gt{w}^{+}_e$ (or both) are sparse, we can expect the denominator $\sum_e(({w^{+}_e})^2 + (\gt{w}^+_e)^2)$ to be small,
which has the effect that the numerator is adaptively weighted higher. 
In this sense, the S{\o}rensen-Dice loss at every pixel $i$ is conditioned on the global image statistics, which is not the case for a Hamming-distance based loss like Binary Cross-Entropy or Mean Squared Error. 

We optimize this loss using the Adam optimizer \cite{kingma2014adam} and additionally condition learning rate decay on 
the Adapted Rand Score \cite{isbi2012challenge} computed on the training set every 100 iterations.
During training, we augment the data set by performing in-plane rotations by multiples of 90 degrees, flips along the X- and Y-axis as well as elastic deformations.
At prediction time, we use test time data augmentation, presenting the network with
seven different versions of the input obtained by a combination of rotations by a multiple of 90 degrees, axis-aligned flips and transpositions.
The network predictions are then inverse-transformed to correspond to the original image, and the results averaged.

\subsection{ISBI Challenge} \label{4_isbi}

The ISBI 2012 EM Segmentation Challenge \cite{isbi2012challenge} is the neuron segmentation challenge 
with the largest number of competing entries.
The challenge data contains two volumes of dimensions 1.5 $\times$ 2 $\times$ 2 microns and 
has a resolution of 50 $\times$ 4 $\times$ 4 nm per pixel. The groundtruth is provided as binary membrane
labels, which can easily be converted to a 2D, but not 3D segmentation. To train a 3D model, we follow the procedure
described in \cite{beier2017multicut}. 
\captionsetup[subfigure]{justification=centering, singlelinecheck=off}
\begin{figure}
\centering
    \begin{subfigure}[t]{0.46 \linewidth}
        \centering
        \includegraphics[width=0.98\textwidth]{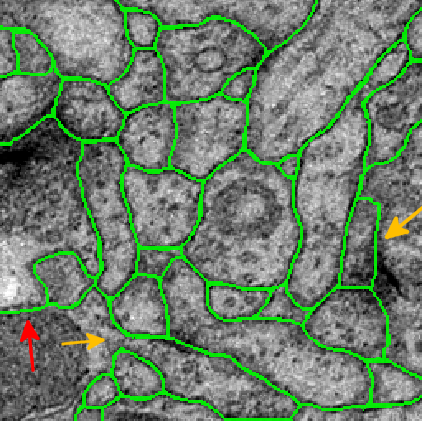}
        \caption{Mutex Watershed} \label{fig:mws1}
    \end{subfigure}\hspace{0.5cm}
    \vspace{0.3cm}
    \begin{subfigure}[t]{0.46 \linewidth}
        \centering
        \includegraphics[width=0.98\textwidth]{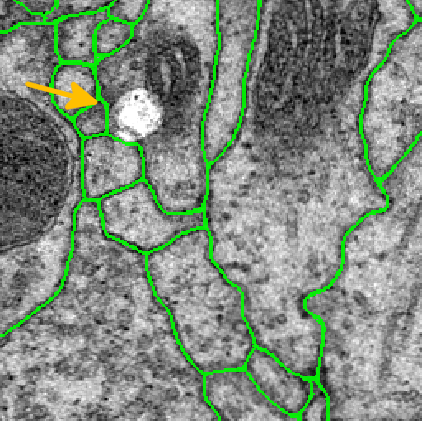}
        \caption{Mutex Watershed} \label{fig:mws2}
    \end{subfigure}
    \vspace{0.3cm}
    \begin{subfigure}[t]{0.46 \linewidth}
        \centering
        \includegraphics[width=0.98\textwidth]{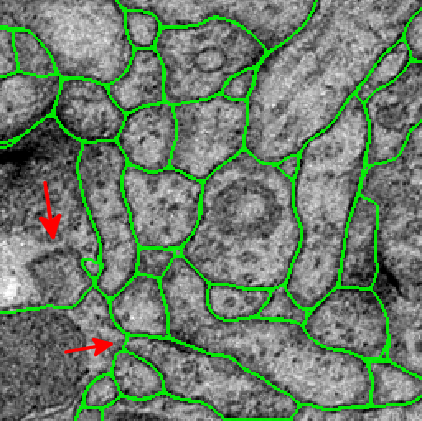}
        \caption{Multicut partitioning based segmentation~(MC-FULL)} \label{fig:mc_full}
    \end{subfigure}\hspace{0.5cm}
    \begin{subfigure}[t]{0.46 \linewidth}
        \centering
        \includegraphics[width=0.98\textwidth]{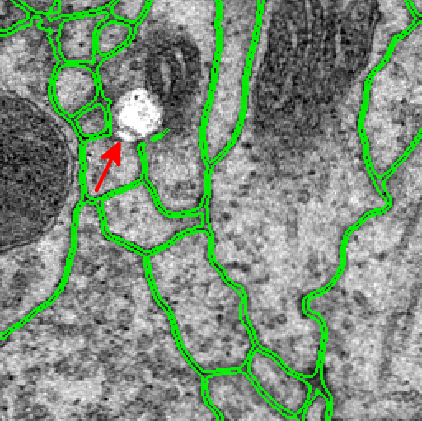}
        \caption{Thresholding of local boundary maps ~(THRESH)} \label{fig:thresh}
    \end{subfigure}%
    
    \begin{subfigure}[t]{0.46 \linewidth}
        \centering
        \includegraphics[width=0.98\textwidth]{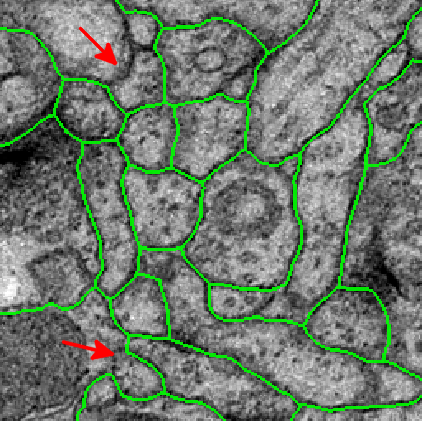}
        \caption{Watershed, seeded at local minima of the smoothed input map~(WS)} \label{fig:ws}
    \end{subfigure}\hspace{0.5cm}%
    \begin{subfigure}[t]{0.46 \linewidth}
        \centering
        \includegraphics[width=0.98\textwidth]{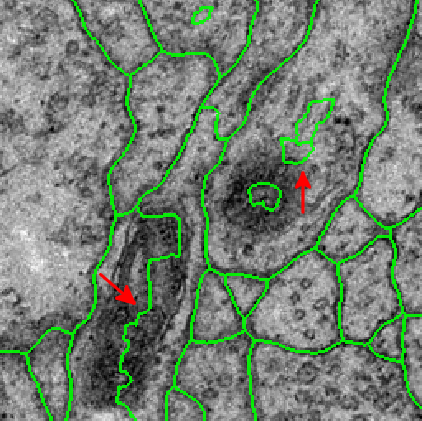}
        \caption{Distance Transform Watershed~(WSDT)} \label{fig:wsdt}
    \end{subfigure}
    \caption{Mutex Watershed and baseline segmentation algorithms applied on the ISBI Challenge test data. Red arrows point out major errors. Orange arrows point to difficult, but correctly segmented regions. All methods share the same input maps.}
    \label{fig:isbi-examples}
\end{figure}%

The test volume has private groundtruth; results can be submitted to the leaderboard.
They are evaluated based on the Adapted Rand Score (Rand-Score) and the Variation of Information Score (VI-Score) \cite{isbi2012challenge}.%

Our method holds the top entry in the challenge's leader board\footnote{{\href{url}{http://brainiac2.mit.edu/isbi\_challenge/leaders-board-new}}} at the time of submission, see Table \ref{tab:isbi-leaderboard}.
This is especially remarkable insofar as it is simpler than the methods holding the other 
top entries. \RED{Three out of four rely on a CNN to predict boundary locations and postprocess
its output with the complex pipeline described in \cite{beier2017multicut}.} 
This post-processing first generates superpixels via distance transform watersheds.
Then it computes a merge cost for local and long-range connections between superpixels.
Based on this, it defines a lifted multicut partioning problem that is solved approximately.
In contrast, our method finds an optimal solution of its objective purely on the pixel level.

\subsubsection*{Comparison with other segmentation methods}
The weights predicted by the CNN described above can be post-processed directly by the Mutex Watershed algorithm. To ensure a fair comparison, we transform the same CNN predictions into a segmentation using basic and state-of-the-art post-processing methods. 
We start from simple thresholding (THRESH) and seeded watershed. Since these cannot take long-range repulsions into account, we generate a boundary map by taking the maximum\footnote{The maximum is chosen to preserve boundaries.} values over the attractive edge channels. Based on this boundary map, we introduce seeds at the local minima (WS) and at the maxima of the smoothed distance transform (WSDT). For both variants, the degree of smoothing was optimized such that each region receives as few seeds as possible, without however causing severe under-segmentation. The performance of these three baseline methods in comparison to Mutex Watershed is summarized in Table~\ref{tab:isbi-baselines}. The methods were applied only in 2D, because the
high degree of anisotropy leads to inferior results when applied in 3D.
In contrast, the Mutex Watershed can be applied in 3D out of the box and yields significantly better
2D segmentation scores.

Qualitatively, we show patches of results in Figure \ref{fig:isbi-examples}.
The major failure case for WS (Figure \ref{fig:ws}) and WSDT (Figure \ref{fig:wsdt})
is over-segmentation caused by over-seeding a region.
The major failure case for THRESH is under-segmentation due to week boundary evidence (see Figure \ref{fig:thresh}).
In contrast, the Mutex Watershed produces a better segmentation, only causing minor over-segmentation (see Figure \ref{fig:mws1}, Figure \ref{fig:mws2}).

Note that, in contrast to most pixel-based postprocessing methods, our algorithm can take long
range predictions into account. To compare with methods which share this property, we turn to the multicut and lifted multicut-based partitioning for neuron segmentations as
introduced in \cite{andres_12_globally} and \cite{horvnakova2017analysis}. As proposed in \cite{andres2012globally}, we compute costs corresponding to edge cuts from the affinities estimated by the CNN via:
\begin{equation}
\label{mc_costs}
    s_e = \begin{cases}
        \log \frac{w^+_e}{1 - w^+_e} , &\text{ if } e \in E^+ \\
        \log \frac{1 - w^-_e}{w^-_e}, & \text{otherwise},
    \end{cases}
\end{equation}
We set up two multicut problems: the first is induced only by the short-range edges (MC-LOCAL), the other by short- and long-range edges together (MC-FULL). Note that the solution to the full connectivity problem can contain ``air bridges'', i.e. 
pixels that are connected only by long-range edges, without a path along the local edges connecting them.
However, we found this not to be a problem in practice.
In addition, we set up a lifted multicut (LMC) problem from the same edge costs.

Both problems are NP-hard, hence it is not feasible to solve them exactly on
large grid graphs. For our experiments, we use the approximate Kernighan Lin \cite{kernighan1970efficient,keuper2015efficient} solver.
Even this allows us to only solve individual 2D problems at a time.
The results for MC-LOCAL and MC-FULL can be found in Table \ref{tab:isbi-baselines}.
The MC-LOCAL approach scores poorly because it under-segments heavily.
This observation emphasizes the importance of incorporating the longer-range edges.
The MC-FULL and LMC approaches perform well. Somewhat surprisingly, the Mutex Watershed yields a better segmentation still,
despite being much cheaper in inference. We note that both MC-FULL, LMC and the Mutex Watershed are evaluated on the same long-range affinity maps (i.e. generated by the same CNN with the same set of weights).

\begin{table}[t]
    \centering
    \begin{subtable}[t!]{0.48\textwidth}\centering
        \begin{tabular}{l c c}
            \toprule
            Method                                  & \hspace{-0.5cm}Rand-Score & VI-Score \\        
            \midrule
            UNet + MWS                              & \textbf{0.98792} & \textbf{0.99183}\\        
            ResNet + LMC \cite{xiao2018deep}        & 0.98788   &  0.99072\\
            SCN + LMC \cite{weiler2017learning}     & 0.98680   &  0.99144\\
            M2FCN-MFA \cite{shen2017multi}         & 0.98383   &  0.98981\\
            FusionNet + LMC \cite{quan2016fusionnet}& 0.98365   &  0.99130\\ 
        \end{tabular}
        \caption{Top five entries at time of submission. Our Mutex Watershed (MWS) is state-of-the-art without relying on the complex lifted multicut postprocessing used by most other top entries.}
        \label{tab:isbi-leaderboard}
    \end{subtable} \quad 
    \par\bigskip
    \begin{subtable}[t!]{0.48\textwidth}\centering
        \begin{tabular}{l c c r}
            \toprule
            Method   & \hspace{-0.5cm}Rand-Score & VI-Score & Time [s] \\        
            \midrule
            MWS      & \textbf{0.98792} & \textbf{0.99183} & 43.3 \\        
            MC-FULL  & 0.98029   & 0.99044 & 9415.8 \\
		    LMC		 & 0.97990   & 0.99007 &  966.0 \\
            THRESH   & 0.91435   & 0.96961 & 0.2 \\
            WSDT     & 0.88336   & 0.96312 & 4.4 \\
            MC-LOCAL & 0.70990   & 0.86874 & 1410.7 \\
            WS       & 0.63958   & 0.89237 & 4.9 \\
        \end{tabular}
        \caption{Comparison to other segmentation strategies, all of which are based on our CNN. Runtimes were measured on a single thread of a Intel Xeon CPU E5-2650 v3 @ 2.30GHz.}
        \label{tab:isbi-baselines}
    \end{subtable}
    \caption{Results on the ISBI 2012 EM Segmentation Challenge.}
    \label{tab:isbi-results}
\end{table}

\section{Conclusion and Discussion} \label{Conclusion}
We have presented a fast algorithm for the clustering of graphs with both attractive and repulsive edges. The ability to consider both
\REVIEW{gives a valid alternative to other popular graph partitioning algorithms that rely on a stopping criterion or seeds.}
 The proposed method has low computational complexity in imitation of its close relative, Kruskal's algorithm. We have shown which objective this algorithm optimizes exactly, and that this objective emerges as a specific case of the multicut objective. \RED{It is possible that recent interesting work \cite{lange2018partial} on partial optimal solutions may open an avenue for an alternative proof.
}

Finally, we have found that the proposed algorithm, when presented with informative edge costs from a good neural network, outperforms all known methods on a competitive bioimage partitioning benchmark, including methods that operate on the very same network predictions.

\section{Acknowledgments}
This work was partially supported by the grants DFG HA 4364/8-1, DFG SFB 1129 from the Deutsche Forschungsgemeinschaft and the Baden-Württemberg Stiftung Elite PostDoc Program.

\ifCLASSOPTIONcaptionsoff
  \newpage
\fi

\bibliographystyle{IEEEtran}
\bibliography{watershed}

\vskip 0pt plus -1fil
\begin{IEEEbiography}[{\includegraphics[width=1in,height=1.25in,clip,keepaspectratio]{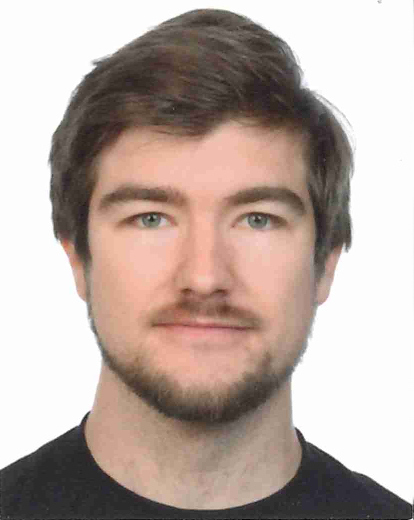}}]%
{Steffen Wolf} is a Ph.D. candidate in the Interdisciplinary Center for Scientific Computing at Heidelberg University. His research interests include 
computer vision, structured prediction and deep learning with applications in image segmentation.
\end{IEEEbiography}\vskip 0pt plus -1fil
\begin{IEEEbiography}[{\includegraphics[width=1in,height=1.25in,clip,keepaspectratio]{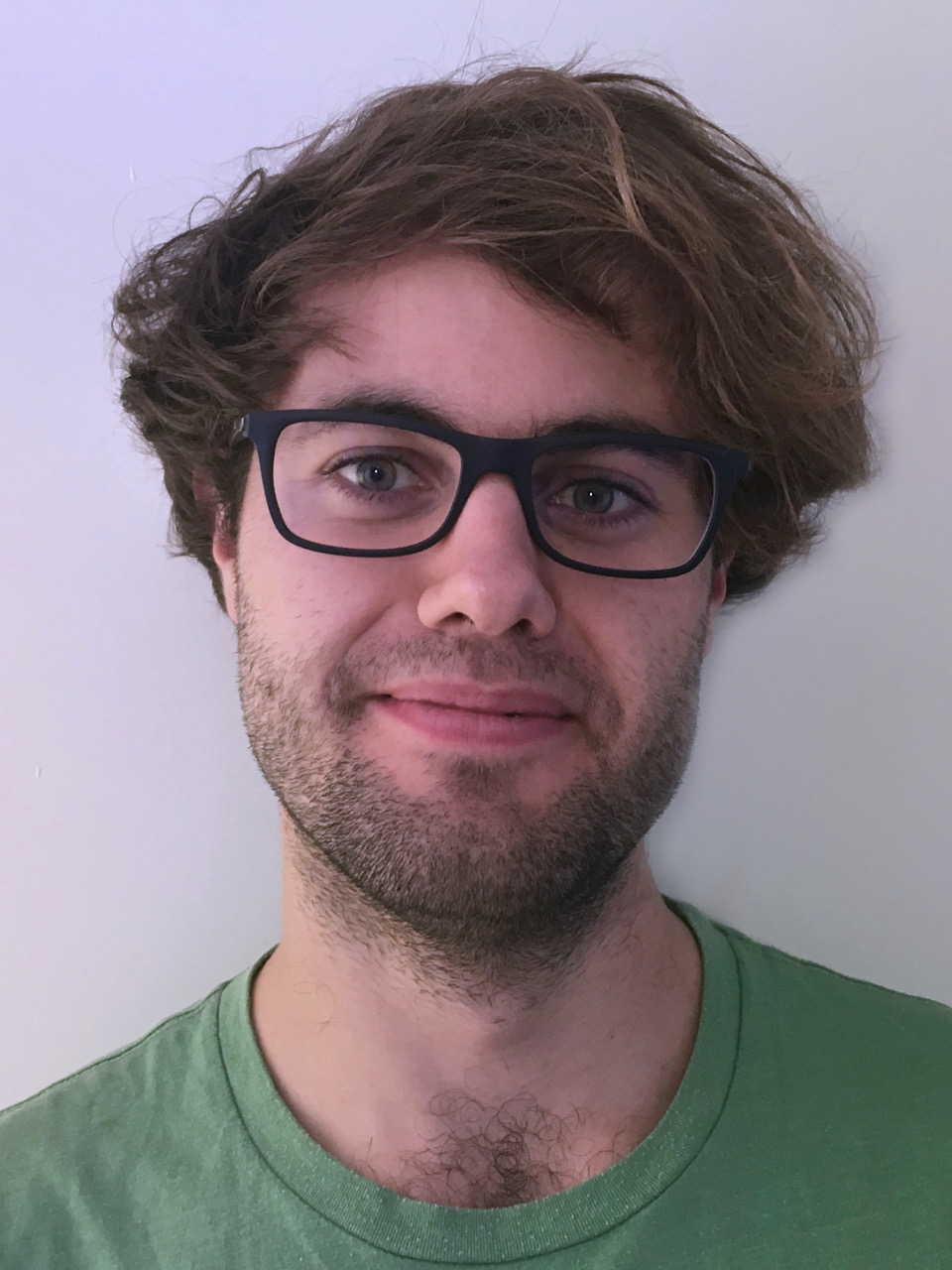}}]%
{Alberto Bailoni} is a Ph.D. candidate in the Interdisciplinary Center for Scientific Computing at Heidelberg University. His research interests include computer vision, image segmentation, deep learning and clustering algorithms, with a focus on their application to automated reconstruction of neural circuit connectivity.
\end{IEEEbiography}\vskip 0pt plus -1fil
\begin{IEEEbiography}[{\includegraphics[width=1in,height=1.25in,clip,keepaspectratio]{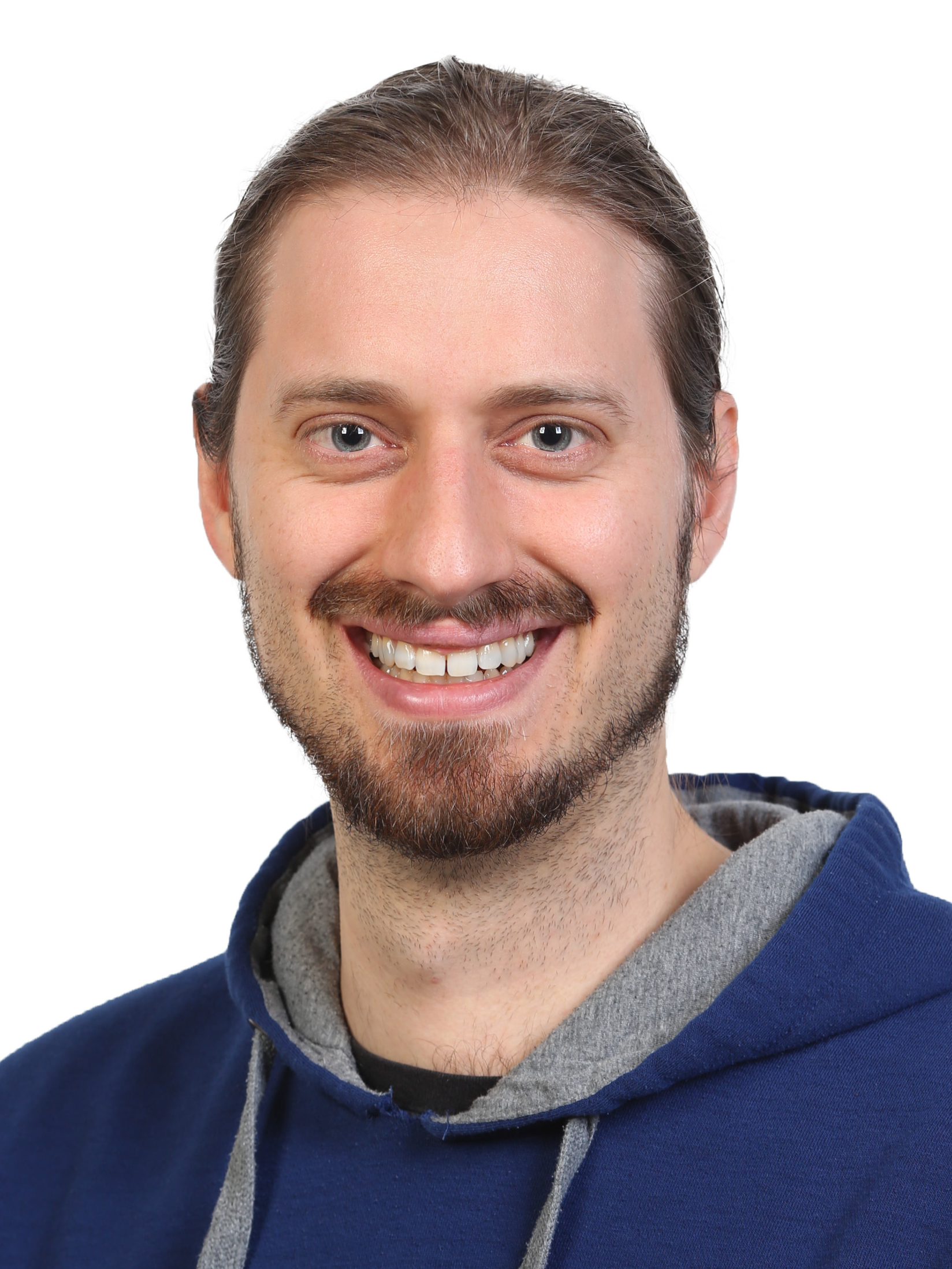}}]%
{Constantin Pape} is a Ph.D. candidate at University of Heidelberg visting the EMBL Heidelberg.
His research interests include biomedical imaging with a focus on deep learning and instance segmentation for large EM datasets.
\end{IEEEbiography}\vskip 0pt plus -1fil
\begin{IEEEbiography}[{\includegraphics[width=1in,height=1.25in,clip,keepaspectratio]{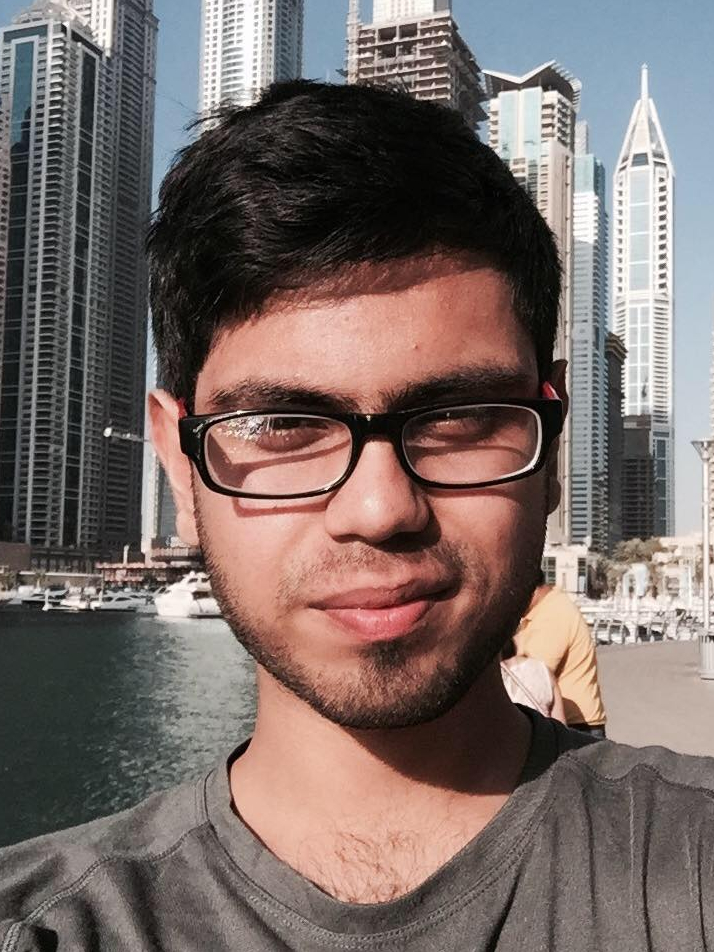}}]%
{Nasim Rahaman} is a MSc student at University of Heidelberg. His research interests are deep learning, reinforcement learning and learning theory.
\end{IEEEbiography}\vskip 0pt plus -1fil
\begin{IEEEbiography}[{\includegraphics[width=1in,height=1.25in,clip,keepaspectratio]{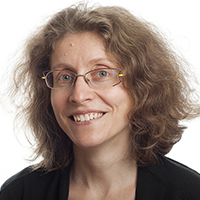}}]%
{Anna Kreshuk} received a diploma in maths from Lomonosov Moscow State University and a PhD in computer science in Heidelberg. She is currently a group leader in EMBL Heidelberg. Her research focuses on automating analysis of microscopy images with machine learning.
\end{IEEEbiography}\vskip 0pt plus -1fil
\begin{IEEEbiography}[{\includegraphics[width=1in,height=1.25in,clip,keepaspectratio]{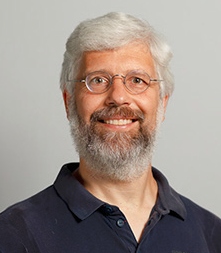}}]%
{Ullrich~K\"othe} received a diploma in physics from the University of Rostock and a PhD in Computer Science from the University of Hamburg. He is currently an associate professor for computer science in the Interdisciplinary Center for Scientific Computing at Heidelberg University. His research focuses on the connection between image analysis and machine learning, and in particular on the interpretability of machine learning results.
\end{IEEEbiography}\vskip 0pt plus -1fil
\begin{IEEEbiography}[{\includegraphics[width=1in,height=1.25in,clip,keepaspectratio]{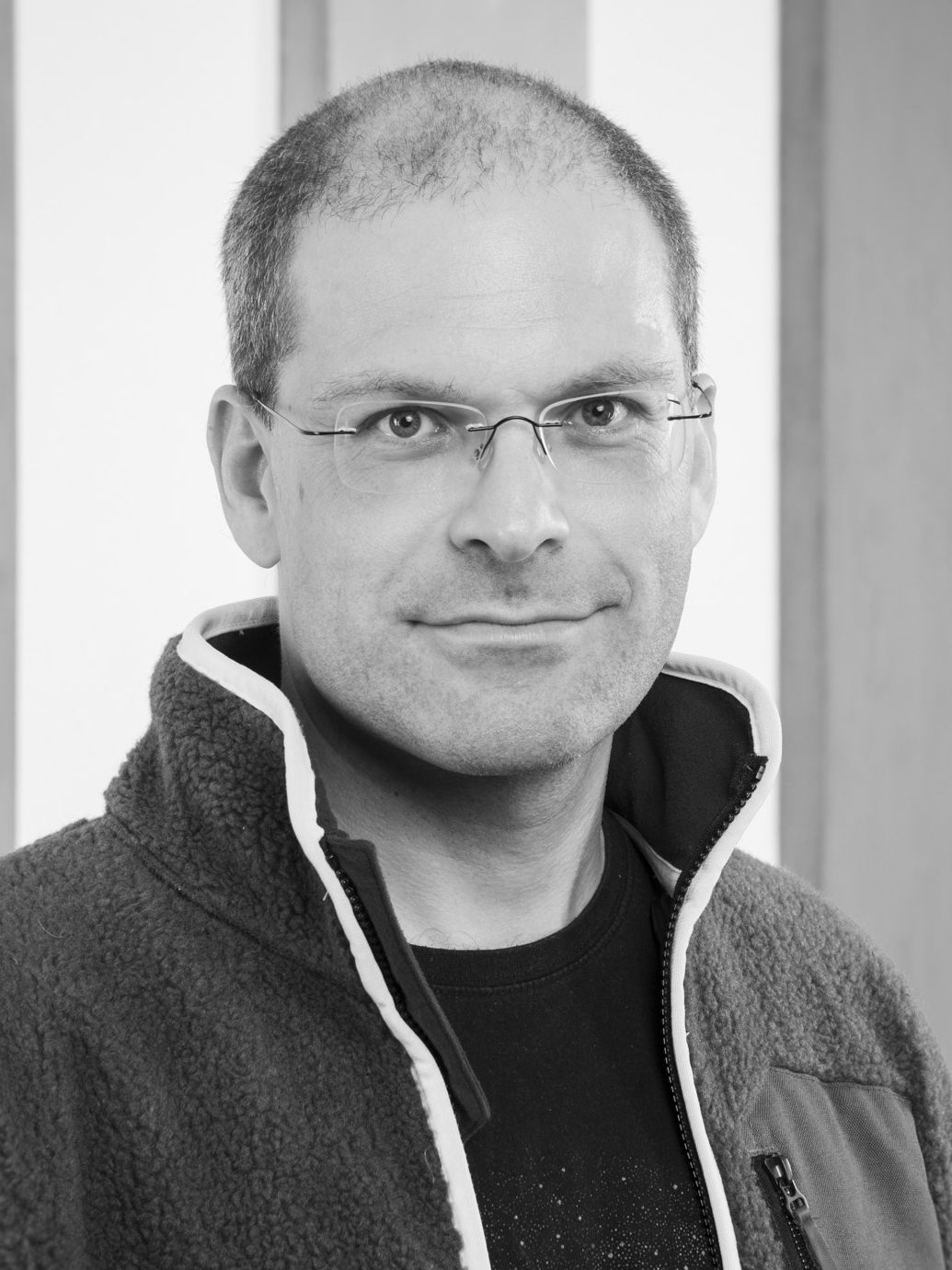}}]%
{Fred A. Hamprecht} is Professor for Image Analysis and Learning at Heidelberg University. His primary research interests are algorithms and how they can be used to solve biological problems. 
\end{IEEEbiography}

\clearpage
\appendices

\section{Property of the minimizers of $Q^p(x)$}\label{sec:proof_minizers}

\renewcommand{\thesection}{4}
\setcounter{theorem}{4}
\boundedsequence*
\begin{proof}
The function $Q^p(x)$ can be explicitly written as (see Eq. \ref{eq:def_Q_p}, \ref{def:scales_MWS} and \ref{eq:def_Q_k_MWS}):
\begin{align}
Q^p(x) =& \sum_{0\leq k< t} \lambda^p_k Q_k (x) \\
= & \, |E| \min_{x' \in \mathsf{\intSC}(\mathcal{G},w)} ||x-x'|| + \sum_{1 \leq k < t} \left | \frac{\tilde{w}_k}{2\tilde{w}_1} \right |^p \sum_{e\in E_{k}} x_e  \\
 = & \, |E| \min_{x' \in \mathsf{\intSC}(\mathcal{G},w)} ||x-x'|| +  \sum_{e \in E} \left | \frac{w_e}{2\tilde{w}_1} \right |^p \,\,x_e. 
\end{align}
We then denote these two terms by:
\begin{align}
Q_\operator{A}^p(x) &:= |E| \min_{x' \in \mathsf{\intSC}(\mathcal{G},w)} ||x-x'||, \\
Q_\operator{B}^p(x) &:= \sum_{e \in E} \left | \frac{w_e}{2\tilde{w}_1} \right |^p \,\,x_e. 
\end{align}

\noindent Intuitively, we now prove that the minimizer $x_p$ of $Q^p(x)$ lies in $\mathsf{\intSC}(\mathcal{G},w)$ by showing that the first term $Q_\operator{A}^p(x)$ is always ``dominant'' as compared to $Q_\operator{B}^p(x)$. \\
First, we note that the gradient of the first term $Q_\operator{A}^p(x)$ has always norm equal to $|E|$ and points in the direction of the closest point $x' \in \mathsf{\intSC}(\mathcal{G},w)$. Given a generic point $y\in \mathbb{R}^{|E|}$, the only two cases when the gradient $\nabla_x\,Q_\operator{A}^p(x)$ does not exists are: i) if $y \in \mathsf{\intSC}(\mathcal{G},w)$; ii) if there are at least two points $x'',x''' \in \mathsf{\intSC}(\mathcal{G},w)$ such that $||y-x''||=||y-x'''||$.
Clearly, $Q_\operator{A}^p(x)$ presents minima only in the first case, when \mbox{$y\in \mathsf{\intSC}(\mathcal{G},w)$}.\\
On the other hand, the second term $Q_\operator{B}^p(x)$ is always differentiable and the norm of its gradient is never greater than $\sqrt{|E|}$:
\begin{equation}
\left|\left|\nabla_x\,Q_\operator{B}^p(x)\right|\right| < \left|\left| \nabla_x \left(\sum_{e \in E} x_e\right) \right|\right| = \sqrt{|E|}
\end{equation}
where we used the fact that $\tilde{w}_k / 2\tilde{w}_1<1$ for every \mbox{$1\leq k < t$}. 
Thus, the magnitude of the gradient given by the first term is always larger compared to the one given by the second term. We then conclude that the objective can always be reduced unless $x_p$ is a point of $\mathsf{\intSC}(\mathcal{G},w)$.
\end{proof}

\end{document}